\documentclass[shortAfour]{sagej}

\usepackage{amsmath}
\usepackage{mathtools}
\usepackage{subcaption}
\usepackage[numbers]{natbib}
\usepackage{graphicx}
\usepackage[table]{xcolor}

\usepackage{enumitem}

\newcommand{\refabsec}[1]{Sec.~\ref{sec:#1}}
\newcommand{\refabsecp}[1]{(Sec.~\ref{sec:#1})}
\newcommand{\refabfig}[1]{Fig.~\ref{fig:#1}}
\newcommand{\refabfigp}[1]{(Fig.~\ref{fig:#1})}

\newcommand{\refabeqp}[1]{(Eq.~\ref{eq:#1})}
\newcommand{\refabtabp}[1]{(Tab.~\ref{tab:#1})}

\begin{document}

\runninghead{Abelha and Guerin}

\title{Transfer of Tool Affordance and Manipulation Cues with 3D Vision Data}

\author{Paulo Abelha and Frank Guerin}
\affiliation{University of Aberdeen}

\corrauth{Paulo Abelha, Department of Computing Science, University of Aberdeen, King's College, AB24 3UE Aberdeen, Scotland.}

\email{p.abelha@abdn.ac.uk}

\begin{abstract}
Future service robots working in human environments, such as kitchens, will face situations where they need to improvise. The usual tool for a given task might not be available and the robot will have to use some substitute tool. 
The robot needs to select an appropriate alternative tool from the candidates available, and also needs to know where to grasp it, how to orient it and what part to use as the end-effector. 
We present a system which takes as input a candidate tool's point cloud and weight, and outputs a score for how effective that tool is for a task, and how to use it.
Our key novelty is in taking a task-driven approach, where the task exerts a top-down influence on how low level vision data is interpreted.
This facilitates the type of `everyday creativity' where an object such as a wine bottle could be used as a rolling pin, because the interpretation of the object is not  fixed in advance, but rather results from the interaction between the bottom-up and top-down pressures at run-time.
The top-down influence is implemented by transfer: prior knowledge of geometric features that make a tool good for a task is used to seek similar features in a candidate tool. 
The prior knowledge is learned by simulating Web models performing the tasks.
We evaluate on a set of fifty household objects and five tasks.
We compare our system with the closest one in the literature and show that we achieve significantly better results.
\end{abstract}


\maketitle

\section{Introduction}


%
%
%

Service robots in everyday environments, such as the home,  will need to deal with a great variety of tools for different tasks. Additionally, in an unconstrained environment they cannot assume that the tool they are accustomed to is always available; they may need to consider substitutes.
We tackle the following problem:
Given a particular task and a set of candidate tool objects, select the best candidate, give it a confidence score and also output where to grasp and how to orient the tool for the task. 
One problem here is intra-class variability, e.g. a new spatula may have quite a different shape to the one we are accustomed to; a second problem is cross-class transfer, e.g. a suitably shaped kitchen knife might be quite effective for lifting pancakes and might be used as a spatula when no other choice is readily available.

We are inspired by the ease with which humans exploit visual and physical similarities in order to find suitable substitutes, when the usual tool is not available. Humans can use a wine bottle to roll dough, or a kitchen knife to lift a piece of cake, or a spatula to cut lasagne. These are everyday creative transfers that come naturally to humans. 
How can we bring robots one step closer to this? 
This problem is a special case of the more general problem faced by robots as they move out of constrained factory settings and into applications in more open environments, where unforeseen situations and objects will be encountered. Open and human environments pose significant challenges for robot manipulation and it has been recognised that ``it is not feasible to preprogram robots for all possible contingencies
they may face'' \cite{Ersen2017}. The problem has been recognised in  Artificial Intelligence (AI) more generally as the `long tail' problem: there is a long tail in the distribution of scenarios in the real world \cite{DavisMarcus2015}; i.e. rare cases, in aggregate,  are very frequent and it is hard to include many of them in training data. Humans seem to handle novel cases using their strong ability to transfer what they know and adapt it appropriately to a new case. Returning to the special case of transfer of tool affordances, this type of transfer, we argue, requires a flexible type of knowledge representation and reasoning.
By knowledge we mean knowledge of what features (e.g. flat surface, concave surface, sharp edge, etc.) and relationships among them make a tool good for a particular task.
By reasoning we mean the judgement of whether a newly presented instance belongs to the set of objects that affords that task; in this case, e.g. does this bottle belong to the set of objects that can roll dough?

\begin{figure*}[t!]
	\centering
	\includegraphics[width=16cm]{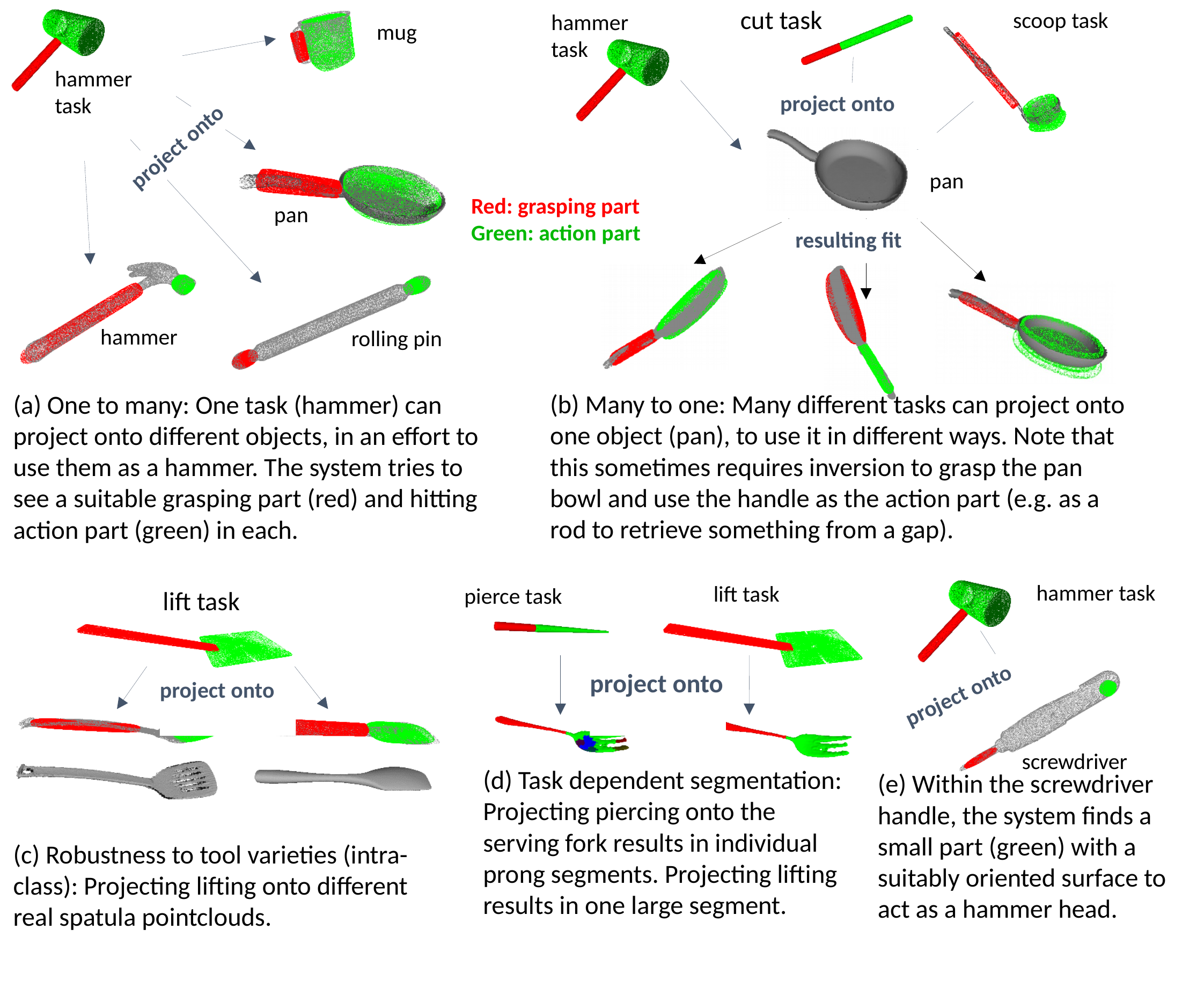}
	\caption{
	Highlights of our ``projection'' technique. Red indicates the grasping part and green the action part.
}\label{keyExamples}
\vspace{-0.5cm}
\end{figure*}

A newly presented object may have many potentially useful features at different places, just as it may have many graspable places.
The judgement of whether or not it belongs to the set of objects good for a task requires a reasoning step at run-time that considers many options to find if there is at least one way it could be grasped and oriented to achieve the task (see for example Fig.~1e).
This is what makes the problem hard.
This reasoning involves interaction between top-down knowledge of what makes a tool good for a task, and bottom-up processing of the features available on the presented object.
This top-down and bottom-up account is inline with knowledge of human visual processing, where it is known that task context can exert a top-down influence on processing \citep{Harel2014a}. 
Top-down and bottom-up processing are also used in some works in computer vision \cite[e.g.][]{HanTPAMI2009}, however the dominant approach in affordance recognition is purely bottom-up \cite[e.g.][]{MyersICRA2015,DehbanICRA2016}.
In pure bottom-up approaches there is a single pipeline incorporating feature recognition and affordance assessment. If we wish to achieve what we called `everyday creative transfers' with a purely bottom-up approach, we would need the training data to provide examples of non-canonical tool-use, such as grasping a tool by the usual end effector and using the handle instead as the end effector. It is difficult for a designer to think of all the alternative grasps and orientations that might be useful at run-time.

Our approach to the problem is inspired by cognitive science research \cite{BipinBook1992} where the same sensor data can be interpreted in alternative ways depending on the top-down pressure being exerted. In our case the task exerts top-down pressure to select the interpretation of a tool, and there are no predefined canonical categories such as spatula, knife, etc.
Borrowing  from the terminology of Indurkhya \cite{BipinBook1992} we call this task-driven interpretation `projection', meaning that the system is projecting what it wants to see in a particular tool when it assesses its effectiveness for a particular task.
Consider the task of hammering, and how this causes four household objects to be interpreted in terms of a grasping part and an action part in Fig.~\ref{keyExamples}a. 
 Under this approach the two problems above (intra-class variation, and cross-class transfer) are not distinct, but are dealt with by the same system. 



Figure~\ref{keyExamples} illustrates several examples of what our system can do, to showcase essential features of the projection approach. 
Our proposed projection process can be seen as a type of reasoning occurring at run-time that considers possible fits between the task requirements and the tools present.
The knowledge of the features and relationships among them, that make a tool good for a task, is learnt during a training process which tries tools in simulation.
 We represent a tool with 25 parameters: tools are composed of two parts, grasping and action (represented as flexible geometric models), and their relationships, along with moment of inertia and mass. This representation also includes the orientation and positioning of the tool for a task. 
We use our training set ToolWeb of 116 mesh models from the Web \citep{AbelhaIROS2017}.
These tools are not just tools for the tasks we are interested in, but tools for a wide variety of kitchen tasks. The training set is upsampled to generate $5,000$ examples.
We converted each of these models to a parameterised tool description.
We then labeled the training set using simulation, and learnt several `task functions' that map from our tool representation to an affordance score for that task. 

At testing time we map a point cloud of a real tool to our tool representation. For this we plant seeds on the candidate point cloud and cut away segments by planting spheres and keeping only the points inside them. We then fit geometrical models to these generated segments and represent the tool with our 25-parameter vector. Each vector then is assessed according to its fitting score and the task function score. The candidate is chosen that has the best fit and task score (with equal weight).
We compare our approach with the closest one in the literature \citep{SchoelerTCDS2016} and show that we achieve significantly better results.

The main contributions of this paper are:

\begin{itemize}
\item A generic framework for representing household tools for use in a variety of tasks
\item A projection algorithm that takes into account the bottom-up 3D data from the candidate tool, and the top-down learned task function (and optionally an ideal tool for a task) in order to output an affordance score and indications of grasping and orientation (tool-pose)
\item A pipeline for building a large dataset of affordance-labeled tools and their usage for different tasks  (by `usage' we mean how to grasp it and orient it to perform the action). Tools are simulated performing tasks in order to determine their effectiveness.
\end{itemize}
This paper extends two previous conference papers \cite{AbelhaIROS2017,AbelhaICRA2016}, the additional material here includes: new experiments which (1) compare results with and without top-down influence from the task; (2) analyse the effect of increasing the number of seeds planted; (3) analyse running times of component algorithms. Also a new task (scooping) has been added, and the explanation of the techniques is more complete than possible within the constraints of previous conference papers.

\section{Related Work}



Here we focus on discussing work in tool affordances for everyday tasks. There are many other approaches that focus on learning affordances for objects, but not necessarily for tool use \citep{Sahin2007};  we are not addressing those here. A recent survey shows that most affordance work has not addressed tool use \citep{JamoneTCDS2016}. In existing approaches there are different examples of output: an affordance binary value - does it afford or not? \citep{Chu2016}; an affordance score, or similarity score between tools \citep{SchoelerTCDS2016,SinapovICDL2008,AgostiniIROS2015}; providing affordance regions, which includes where the robot could grasp a tool \citep{MyersICRA2015}; given a certain grasping, knowing how the orientation of the tool affects the affordance \citep{MarICHR2015}; detecting objects parts to be used as knowledge primitives in robot planning \citep{TenorthIROS2013}. 

We want the robot to be able not only to assess a tool, but also know how to use it properly for the given affordance. We can think of a spectrum, where the left end means outputting a binary value for an affordance and the right end, outputting a graded score plus a full motor program for grasping and using the tool for a given task (considering things such as the expected torque by assessing the tool's inertia). To the best of our knowledge no work in affordance of tools has been done that falls all the way to the right end of the spectrum. Nonetheless we fall further to the right than any other approach since we output: the affordance score; a fitted geometric model for the grasping part; a transformation that tells the robot, given the grasping position, how to orient and use the assessed tool for a known task  (which we call the `usage'); and the expected moment of inertia for the tool given the usage. 

Regarding the data used, most approaches work with RGB-D \citep{MyersICRA2015,DehbanICRA2016}; sometimes just color information \citep[using robot NAO's camera]{Wang2014a}; or even RGB-D videos \citep{Koppula2012}. Some approaches, like us, work with full point clouds \citep{SchoelerTCDS2016}. Our approach could be integrated in a real robotics setup where the robot can grasp the object and rotate it to get a full point cloud \citep{KraininIJRR2011,KraftTCDS2010,WelkeICRA2010,WuACMTG2014}. Among existing work the most sophisticated we found explicitly considers the parts of a tool and the relation between them \citep{SchoelerTCDS2016}, some others also consider parts but are not focused on tools \cite{Biegelbauer2008,KarthikICPR2014,UgurCSNO2011,KroemerICRA2012}), while some other works do not consider part relations \citep{MyersICRA2015,DehbanICRA2016}. For tasks such as hammering or lifting a pancake it is not enough to recognise a suitable head, it is also required that there is a handle at a suitable distance and orientation relative to the head.

Existing work tends to approach the labeling of a tool in a binary way; i.e. a tool affords a certain task or it does not \citep{MyersICRA2015,SchoelerTCDS2016}. 
This `all or nothing' labeling can make it difficult for the learning system to figure out how the  various parts of a tool, and their relationships, contribute to the success.
For example, consider the effectiveness of a range of tools for hammering. A hammer gets gradually less effective as the handle gets shorter and the head moves closer to the grasping place. By having a real number value to label the effectiveness we could learn how effectiveness is a function of this length.

Furthermore the training sets in existing work are usually limited by the need to hand-label many examples \citep{MyersICRA2015}, or to generate synthetic models of tools in a particular class, according to a recipe \citep{SchoelerTCDS2016}. Tools generated by a recipe tend to be quite similar to each other, whereas real world tools exhibit more variation. Even carefully selected training examples are likely to miss some unusual uses of objects. The resulting trained classifiers are good at reporting similarity to the training set, where the training set is usually a set of `in category' tools; they do not capture more generally how variations in relationships between parts and different usages relate to performance on a particular task. We need a system that learns a more fine grained understanding of how the parts of a tool (and their relationships)  and its usage relate to performance on a task.

Current pure data-driven (e.g. deep learning) approaches have been very successful in learning from large datasets \citep{LeCun2015,Schmidhuber2015}, including 
computer vision with 3D data \citep{MahlerDexNet2017,LenzIJRR2015,Savva2015}. 
It remains to be seen how far can we push these approaches to generalise well to everyday task situations \citep{LakeBBS2017}. In \cite{Levine2015} the authors make an interesting case for learning end-to-end visuomotor skills in a real robot for everyday tasks. 
Potentially such an approach could learn for itself the important features (e.g. parameters of shape and relationship between parts) which determine the affordances and best usage of any unknown candidate tool.
However, in order to flexibly recognise those features across varied test cases, the system would need to be able to internally isolate the features from other aspects of how they appeared in training examples, e.g.  the relationships the features occurred in. 
The relationships among features in test cases may be quite different to the training cases. 
Unfortunately this seems to be a problem in deep learning, which still finds it challenging to `disentangle' \cite{BengioRepLearn2015} the factors that explain the data (i.e. the tool features and their relationships in this case).
We believe the possible tools and situations that a robot might encounter in open scenarios are too varied to be learned without some sort of knowledge transfer, e.g. knowledge of basic naive physics, such as a child might learn during development. 
Although deep learning has conquered grasping \citep{MahlerDexNet2017}, the degrees of freedom to be considered in manipulation scenarios involving tools acting on other objects are much higher.
Recent work by Lake et al. \cite{LakeBBS2017} is also sceptical of the extent to which deep learning alone can solve open real world problems, and argues for the inclusion of such things as 
causal models, intuitive physics, and compositionality.


To the best of our knowledge there are no available large datasets on 3D sensor data of tools and their corresponding affordance, part (grasp and action) and usage labels for everyday tasks. The closest example we found was Myers et al. \cite{MyersICRA2015}, where there is a large dataset of RGB-D data with per-pixel affordance labels. However there is only part labeling, e.g. of a graspable part, or an action part such as scoop or cut. There is no indication of the orientation in which a tool should be used, and which edge or surface should contact the material to be acted on. We attempted but failed to extract full point clouds from this data, probably because the tools rotate on a surface while the background does not (making it difficult for registration, from frame to frame), and also the rotation is not at a uniform speed.
Secondly there is the work of Bu et al. \cite{Bu2014}, where the authors present a large dataset of CAD models (mesh surfaces) gathered from the Web with their corresponding category labels. However there are no labels for usage, affordances nor weight.
Furthermore the problem requires that we also consider different usages for the same tool, i.e., grasping and orienting in a different way for a different task. Being able to learn both affordance and usage in a pure data-driven approach remains thus an open question due to the difficulties involved in gathering the dataset and labelling it. Our work with simulation could be used to help to label such a dataset.

\section{System Description}
\label{sec:system}

\subsection{Overview}

\begin{figure*}[t!]
	\centering
	\includegraphics[width=15cm]{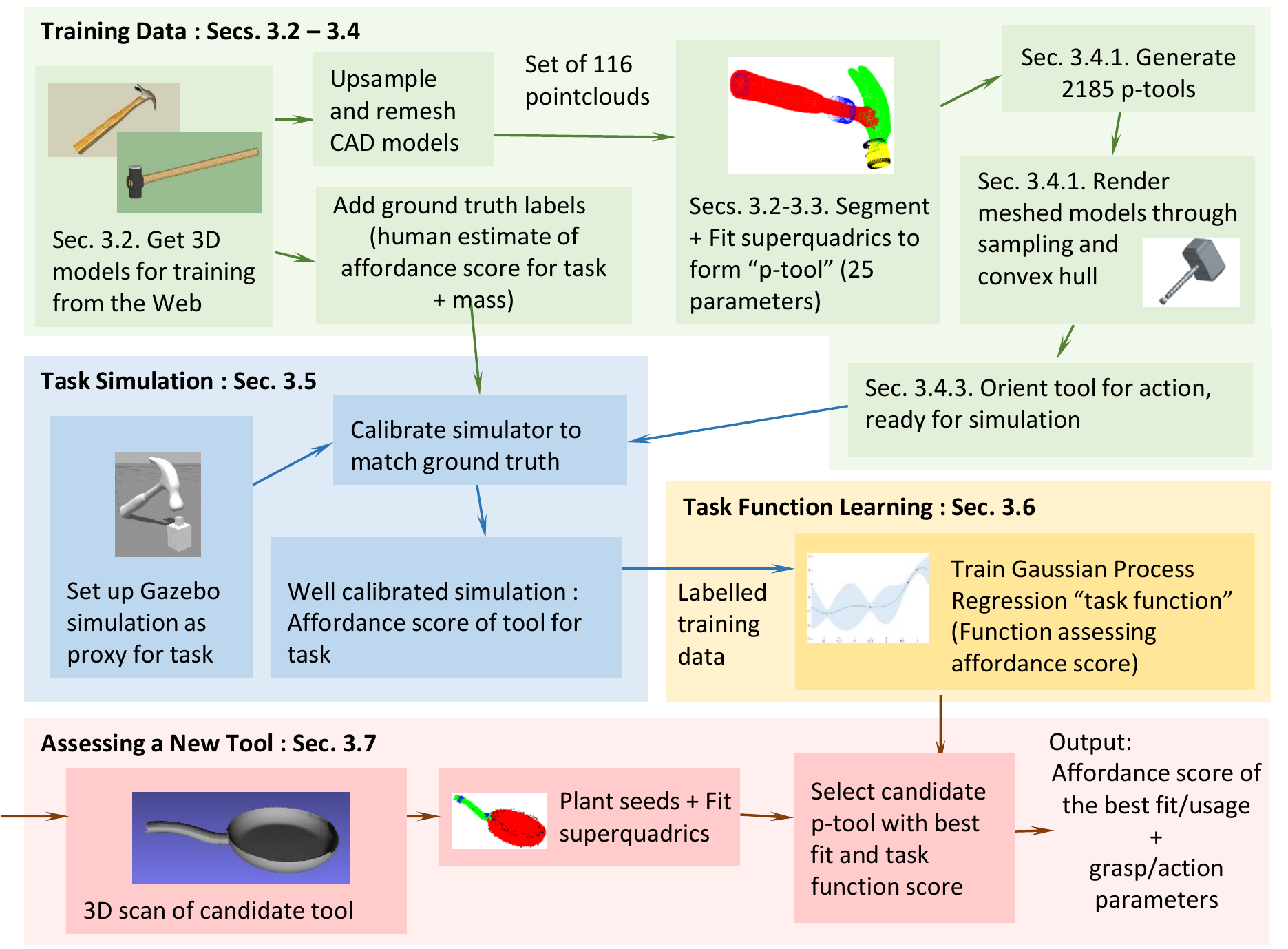}
	\caption{Overview of the system; how it is trained and used.}
	\label{fig:overview_system}
	\vspace{-1ex}
\end{figure*}

The system gets as input a task, a point cloud, and tool weight. It outputs an affordance score for the task; two regions of the point cloud, one for grasping and another for the action part (end-effector) \refabsecp{projection}; and positioning and rotation parameters to position the tool relative to the task, for instance, the nail's centre for hammering a nail or the pancake's centre for lifting pancake \refabsecp{ptool_positioning}.

There are four main parts to our approach (coloured in \refabfig{overview_system}): preparing training data; simulating a task; learning a task function; and assessing a new tool. These parts and subparts have corresponding sections below, indicated in \refabfig{overview_system}.

\subsection{Training Data}
\label{sec:training_data}

\subsubsection{ToolWeb Dataset}
\label{sec:toolweb_dataset}

\begin{figure}[h]	
	\centering	
	\includegraphics[width=8cm]{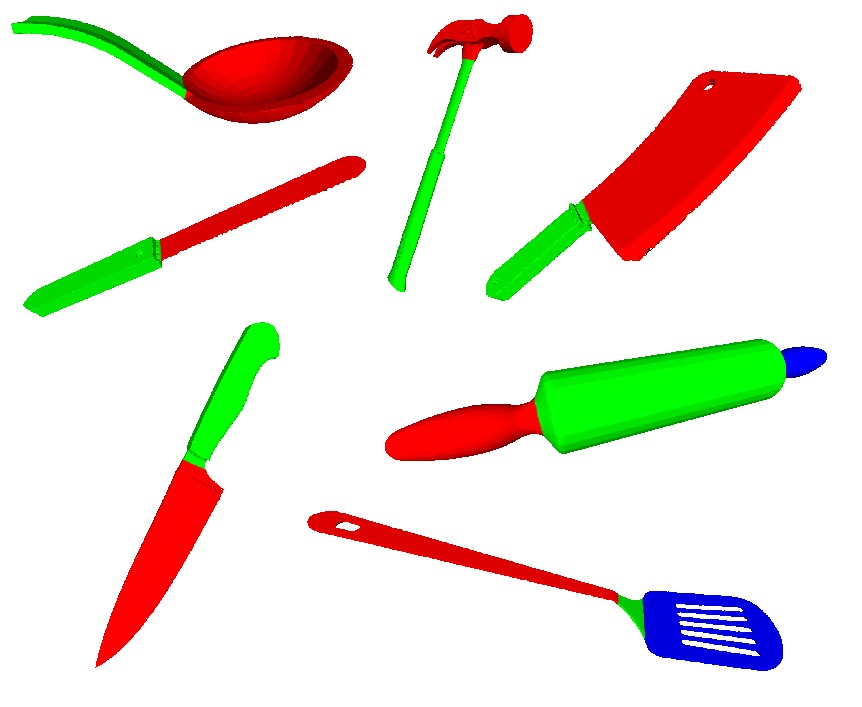}
	\caption{ToolWeb dataset examples. Different colours represent different segments (colours here do not code any grasp or action part). Point clouds in ToolWeb are automatically up-sampled, re-meshed, re-scaled and segmented. Re-printed from \cite{AbelhaIROS2017}.}
	\label{fig:ToolWeb}
\end{figure}

We downloaded 116 CAD models from the web (3DWarehouse\footnote{https://3dwarehouse.sketchup.com}) corresponding to everyday kitchen objects \refabfigp{ToolWeb}. The CAD models had only a handful of points so we needed to upsample the points and reconstruct the mesh. We set up an automatic pipeline to run for every CAD model (made possible because of MeshLab Server application):

\begin{enumerate}
\item Upsample the points via Ball-Pivoting in Meshlab (default settings)
\item Calculate the normals for the points via the Matlab function \textit{pcnormals} in the Computer Vision Toolbox \footnote{http://uk.mathworks.com/help/vision/ref/pcnormals.html}(normals are necessary for the automatic segmentation  Sec.~\ref{sec:automatic_segmentation}).
\item Recalculate the mesh using Poisson Reconstruction in MeshLab (default settings)
\end{enumerate}

We also require the mass for every tool. For this a human hand-labeled, for every task, each tool with an affordance score and mass value according to the usual material for that tool and its size. The affordance score label is a category variable defining four categories for every task: 1 - no good; 2 - slightly effective; 3 - good with effort; 4 - very good. This labeling is the only non-automatic step in our framework and there could be ways to automate it by knowing the material of the object and the volume of the point cloud. We name the remeshed and human-labelled dataset \textit{ToolWeb}.

\subsubsection{Automatic Segmentation}
\label{sec:automatic_segmentation}

We segment each mesh using Triangulated Surface Mesh Segmentation (TSMS) \citep{YazTSMSegmentation2012}\footnote{https://doc.cgal.org/latest/Surface\_mesh\_segmentation}.
TSMS uses the Shape Diameter Function (SDF) \citep{ShapiraSDF2008} as a measure of local object thickness for each facet and then applies a soft-clustering on the SDF values for each facet, followed by a hard clustering and graph-cut algorithm that considers three aspects: soft clustering values,  dihedral-angle between facets and concavity (if the angle is concave it is weighted more than if its convex). There are two hyper-parameters for TSMS: the number of clusters $k$ for the soft clustering; and a smoothing parameter $\lambda$ that penalises large number of segments.

For each mesh we generated $25$ segmentation options by iterating over $k$ from $3$ to $7$ in steps of $1$ and $\lambda$ from $0.1$ to $0.5$ in steps of $0.1$ \refabfigp{segm_options}. This may generate options with very small segments (number of points in a segment is less than 5\% of the overall number of points in the point cloud). From each of the $25$ options we eliminate small segments by merging each of its point with the closest segment to it; we repeat this until we have no small segments. It is possible for meshes to have only one segment, which is necessary for objects like the bowl or plate. We select only one segmentation option for each mesh by fitting superquadrics to the segments and choosing the segmentation option with the best fit score averaged over all segments.

\subsection{Superquadrics}
\label{sec:superquadrics}

The term superquadrics was introduced by \cite{Barr1981} and it refers to a general family of shapes that includes superellipsoids, superhyperboloids and supertoroids. We don't make use of superhyperboloids or supertoroids, but we introduce another shape, the superparaboloid, giving our own formulation for it to model, for instance, bowls, or spoon and ladle heads \refabfigp{ptools}. Here we use superquadrics to refer to both superellipsoids and superparaboloids.

Superquadrics offer a flexible and concise way of representing a number of different shapes. We use them to represent each of the two parts of a tool: grasp or action. With superquadrics it is possible to cast the problem of recovering a superquadric from a point cloud as a minimization problem: given a point cloud we can efficiently recover an approximate superquadric representation. For detailed explanation and formulation we refer the reader to \cite{JacklicSQBook,AbelhaSQ2017}.

\begin{figure}[t!] 
  \centering
  \begin{subfigure}[b]{0.5\linewidth}
    \centering
    \includegraphics[width=0.85\linewidth]{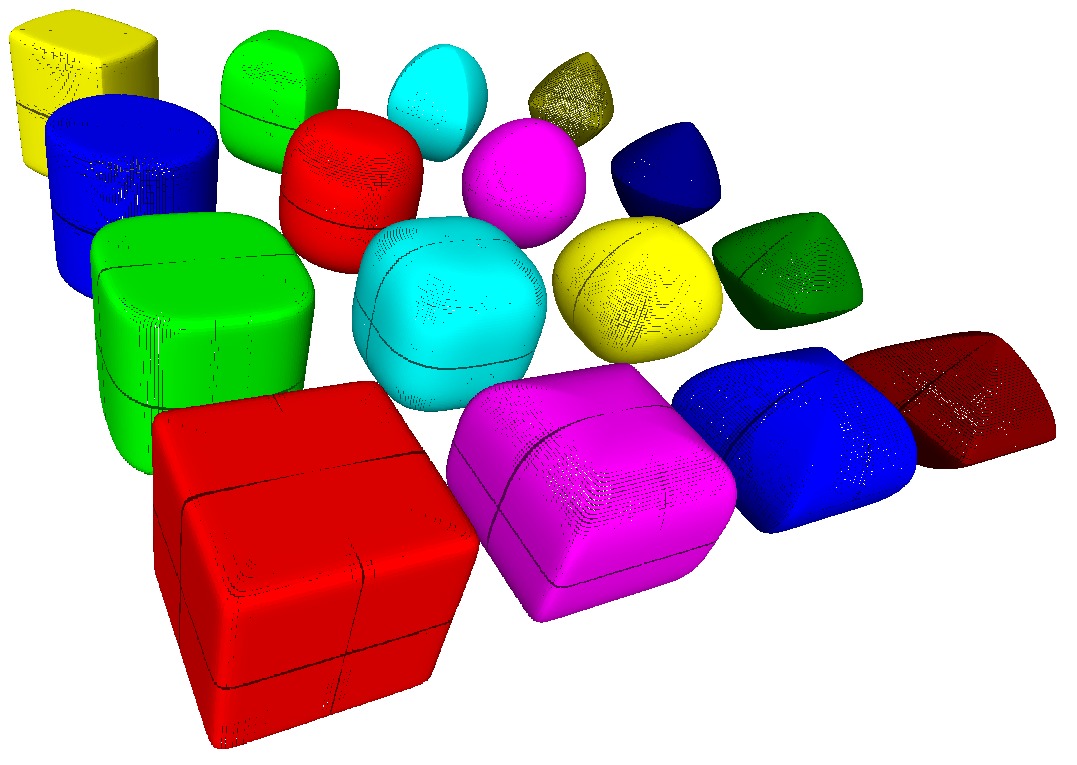} 
    \caption{Superellipsoids} 
    \label{fig:superellipsoids} 
    \vspace{2ex}
  \end{subfigure}
  \begin{subfigure}[b]{0.5\linewidth}
    \centering
    \includegraphics[width=0.85\linewidth]{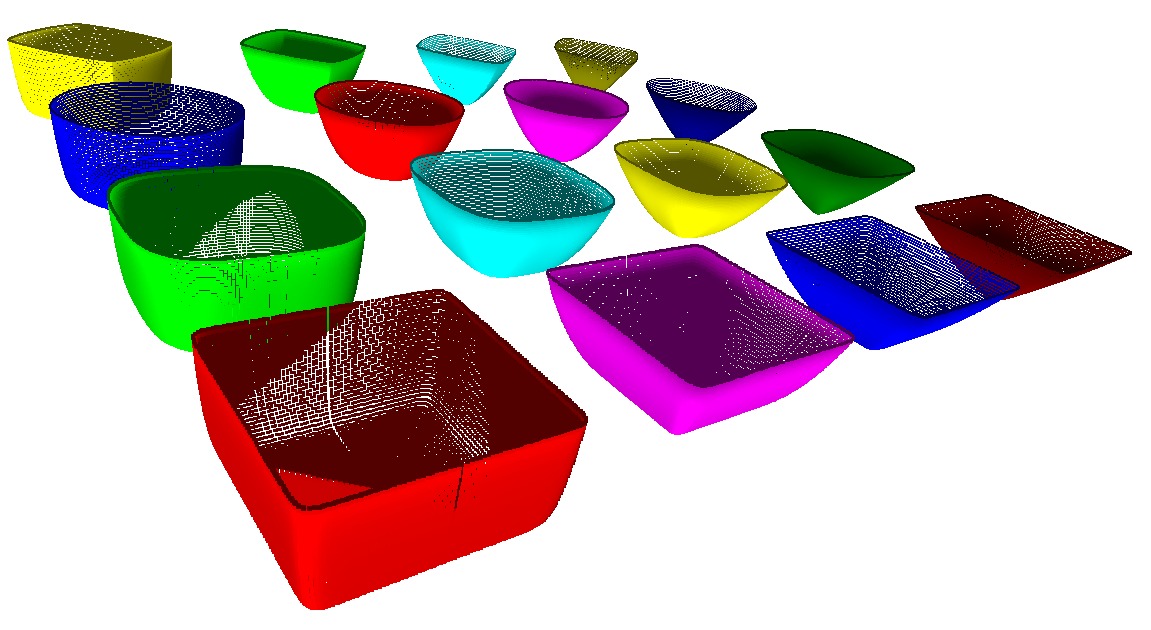} 
    \caption{Superparaboloids} 
    \label{fig:superparaboloids} 
    \vspace{2ex}
  \end{subfigure} 
  \caption{Examples of superquadrics; each axis is varying one of the two shape parameters from $0.1$ to $2.0$ and the three scales parameters are fixed to be the same.}
  \label{fig:intro_superquadrics}
\end{figure}

\subsubsection{Superellipsoids}

Superellipsoids are obtained by the spherical product of two superellipses to obtain a 3D surface \citep{Barr1981} with the implicit equation

\begin{align*}
	\bigg( \bigg( \frac{x}{a_1}\bigg) ^\frac{2}{\epsilon_2} + \bigg( \frac{y}{a_2}\bigg) ^\frac{2}{\epsilon_1} \bigg) ^\frac{\epsilon_2}{\epsilon_1} + \bigg( \frac{z}{a_3}\bigg) ^\frac{2}{\epsilon_1} = 1
\end{align*}

where parameters $a_1$, $a_2$ and $a_3$ define the size of the superellipsoid in the $x$, $y$ and $z$ dimensions respectively; and  $\epsilon_1$ and $\epsilon_2$ control the shape (see Fig. \ref{fig:superellipsoids}). Note that by setting $a_1 = a_2 = a_3 = 1$ and $\epsilon_1 = \epsilon_2 = 1$ we get the unit sphere. We can then build the parametric function

\begin{align*}
	F(\mathbf{x};\Lambda) = \bigg( \bigg( \frac{x}{a_1}\bigg) ^\frac{2}{\epsilon_2} + \bigg( \frac{y}{a_2}\bigg) ^\frac{2}{\epsilon_1} \bigg) ^\frac{\epsilon_2}{\epsilon_1} + \bigg( \frac{z}{a_3}\bigg) ^\frac{2}{\epsilon_1}
\end{align*}

where $\mathbf{x}$ is the point $\mathbf{x} = (x,y,z)$ and $\Lambda = (a_1, a_2,a_3,\epsilon_1, \epsilon_2)$ contains the parameters. The function above is called the \textit{inside-outside} function because it provides a way to tell if a point $\mathbf{x}$ is inside~($F<1$), on the surface~($F=1$) or outside~($F>1$) the superellipsoid.

It is also possible to extend $\Lambda$ and define a superellipsoid in general position and orientation in space. We use $3$ extra parameters $(px,py,pz)$ for the position of its central point, and $3$ more $(\theta, \phi, \psi)$  for the $ZYZ$ Euler angles that define its orientation as a series of three rotations: a rotation of $\theta$ around the $z$ axis, followed by a rotation of $\phi$ around the $y$ axis and a final rotation of $\psi$ around the $z$ axis. Now we have $F(\mathbf{x};\Lambda)$, with $\Lambda = (a1, a2,a3,\epsilon_1,\epsilon_2,\theta,\phi,\psi\,px,py,pz)$ and are able to define a great variety of shapes and sizes in general position and orientation with $11$ parameters.

\subsubsection{Superparaboloids}
We introduce our formulation of the superparaboloid, with the following inside-outside function:

\begin{align*}
F(\mathbf{x};\Lambda) = \bigg( \bigg( \frac{x}{a_1}\bigg) ^\frac{2}{\epsilon_2} + \bigg( \frac{y}{a_2}\bigg) ^\frac{2}{\epsilon_2} \bigg) ^\frac{\epsilon_2}{\epsilon_1} - \bigg( \frac{z}{a_3}\bigg)
\end{align*}

The superparaboloid $\Lambda$ parameter set is analogous to the one for superellipsoids, also containing $11$ parameters.

\subsubsection{Deformations}
In our approach we include two known extension to superellispoids: \textit{tapering} and \textit{bending} deformations. We use the tapering deformation introduced in \cite{JacklicSQBook} that linearly thins or expands the superquadric along its $z$ axis. Tapering requires two extra parameters $K_x$ and $K_y$ for tapering in the $x$ and $y$ directions. For the bending deformation, we use the circle function to deform the superellispoid and bend it positively on $X$ along $Z$. There is only one parameter $k$ defining the bending circle's radius. Finally, we arrive at $\Lambda = (a_1,a_2,a_3,\epsilon_1,\epsilon_2,\theta,\phi,\psi\,K_x,K_y,k,px,py,pz)$ being our final set of $14$ parameters to define a superquadric tapered in general position and orientation. For our current approach the tapering and bending parameters are ignored when the superquadric is a superparaboloid.


\subsubsection{Superquadric Fitting}
\label{sec:superquadric_fitting}

\begin{figure}[h]		
	\centering
	\includegraphics[width=8cm]{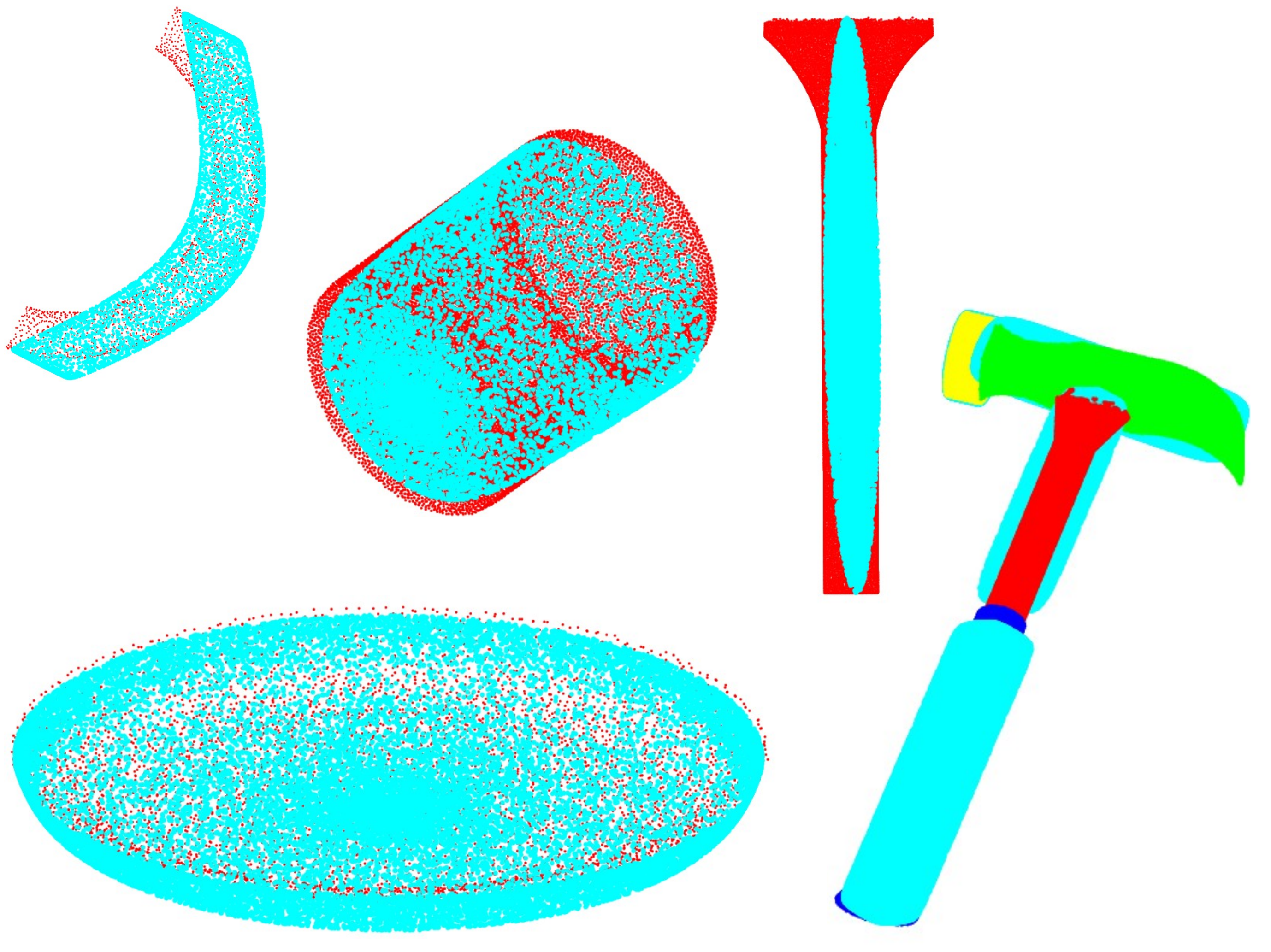}	
	\caption{
	Superquadrics (cyan) fitted to various point clouds (other colours). We show fits for a mug's handle and cup segments; a spatula's handle; an entire hammer and its segments. Note that some fits represent very accurately the point cloud while others not quite so (spatula's handle) or the complete bending of the mug's handle.}
	\label{fig:SQ_fits}	
\end{figure}

We use \textit{superquadric fitting} to recover a superquadric from a point cloud \refabfigp{SQ_fits}. Since we have an inside-outside function $F(\mathbf{x};\Lambda)$ for checking whether a point is inside, on the surface or outside the superquadric, we can cast the task of recovering a superquadric from a point cloud as an optimisation problem \citep{JacklicSQBook}. Given a point cloud as a set of $N$ 3D points we want the values for the parameters in $\Lambda$ that minimise the algebraic distance from the points to the superquadric model. That is, we want the parameter values for which most of the points lie on the superquadric's surface ($F(\mathbf{x};\Lambda)=1$). We cast this as the as a minimisation problem and add an exponent and a term to the minimisation terms for better performance \citep[Chapter 4]{JacklicSQBook}, arriving at

\begin{align*}
\min_{\Lambda} \sum_{i=1}^{N} (\sqrt{\Lambda_1\Lambda_2\Lambda_3}F^\epsilon_1(\mathbf{x};\Lambda)-1)^2
\end{align*}

We use the Levenberg-Marquadt method for nonlinear least squares minimisation. The fitting procedure can be performed on any point cloud; in our work, we use it to fit superquadrics to segmented point clouds \refabsecp{ptool_extraction} and also to point clouds cut out from the original point cloud by planting seeds on it \refabsecp{projection}. Segments and cut-out point clouds are point clouds themselves hence below we refer to fitting as always being done in a point cloud.

When fitting we:

\begin{itemize}
\item Downsample the point cloud to $1000$ points, which we empirically discovered to be a good trade-off between keeping the point cloud's shape and having a fast fitting
\item Initialise the superquadric scale parameters to the bounding box of the point cloud 
\item Initialise the superquadric shape parameters as a ``cylinder'', i.e., $\epsilon_1=0.1$ and $\epsilon_2=1$.
\item Initialise the orientation, tapering and bending parameters to $0$
\item Initialise the superquadric's central position to be the point cloud's mean in the $x$, $y$, and $z$ axes
\end{itemize}

We leave all parameters to be optimised within the boundaries

\begin{equation}
\label{eq:opt_bound}
  \begin{aligned}
\Lambda_{min}=(\mathbf{b},0.1,0.1,0,0,0,-1,-1,c_3,\mathbf{q}) \\
\Lambda_{max}=(\mathbf{c},2.0,2.0,\pi,\pi,\pi,-1,-1,1,\mathbf{r})
  \end{aligned}
\end{equation}

The the minimum $\mathbf{b}$ and maximum $\mathbf{c}$ sizes are set to be, respectively, $80\%$ and $120\%$ of the point cloud's bounding box. The shape parameters are between $0.1$ and $2$. The orientation parameters are between $0$ and $\pi$. The tapering parameters are between $-1$ and $1$ for maximum tapering in both directions. The bending parameters is between the maximum scale in $z$ for maximum bending ($c_3$) and $1$ for minimum bending. The position boundaries $\mathbf{q}$ and $\mathbf{r}$ are set to be, respectively, the minimum and maximum value of the point cloud in the three dimension.

The fitting process is split into four sub-fittings that we call: `normal superellipsoid', `tapered superellipsoid', `bent superellipsoid' and `superparaboloid' fitting. We do this because: 1) the superparaboloid has a different inside-outside function; 2) we discovered, for superellipsoids, that trying to optimise tapering and bending along with the other parameters did not always lead to a good fit. The normal superellipsoid fitting is done by removing the tapering and bent parameters from the optimisation. The tapered superellipsoid fitting is performed by removing the bending parameters. The bent superellipsoids fitting is done by removing tapering parameters. The superparaboloid fitting is done without tapering or bending parameters and with the same optimisation boundaries \refabeqp{opt_bound}.

\subsubsection{Selecting the superquadric}
\label{sec:selecting_SQ}

In order to choose among the four fitted superquadric options, we choose the superquadric that minimises the point-wise distance between the point cloud $P$ and the superquadric's point-sampled point cloud $S$. We obtain the superquadric's point cloud $S$ using our close-to-uniform sampling method \citep{AbelhaSQ2017} to generate $1000$ points (the same number of points as $P$).

The distance $D(P,S)$ between closest points of $P$ and $S$ is not necessarily symmetric, it might be that $D(P,S) \neq D(S,P)$. For example, a small superquadric point cloud $S$ can be fitted to a large point cloud $P$ and have its points relate to close neighbours in $P$, but there may be points in $P$ that are distant from any point in $S$; in this sense, we can say $S$ is well fitted to $P$, not the reverse. We want both $P$ and $S$ to be well-fitted to the other.  We calculate $D(P,S)$ and $D(P,S)$ for all four superquadrics' point clouds  and select the superquadric that minimises $D(P,S)+D(P,S)$.

%

\subsection{Tool Representation: the p-tool}
\label{sec:ptool}

\subsubsection{Overview}

\begin{figure}[h]		
	\centering
	\includegraphics[width=6cm]{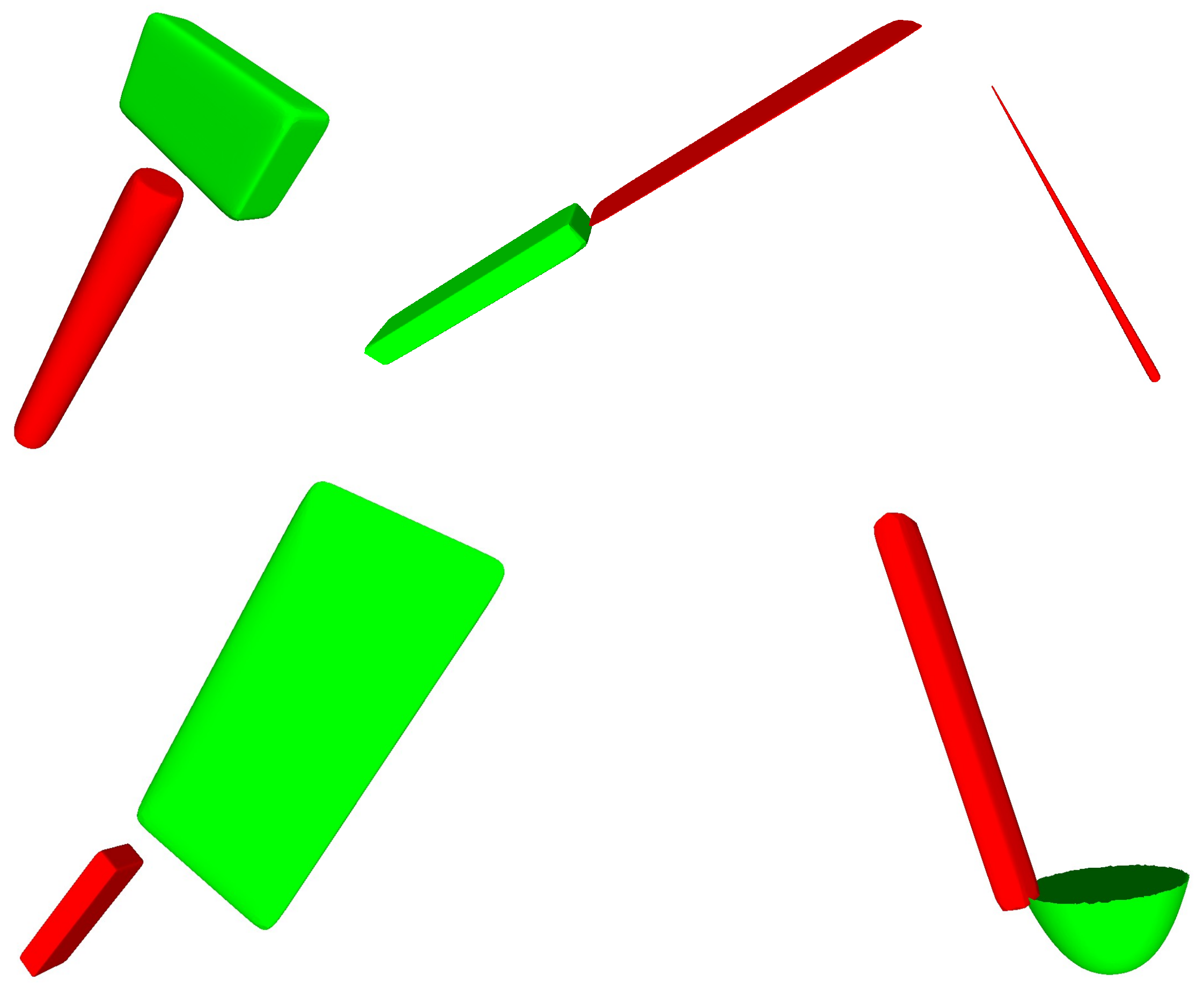}	
	\caption{Different p-tools: red and green represent different segments than can be used as grasp or action part. We can map a point cloud to one or more p-tools and, conversely, render a p-tool into a point cloud/mesh. Here shown rendered p-tools for: hammer; bread knife; chopstick; Chinese knife; ladle.}
	\label{fig:ptools}	
\end{figure}

We represent a tool as a $25$-dimensional vector containing information about the grasp and action part, relationship between these parts and mass. This provides a flexible and generic way of representing many different tools used for a variety of tasks (see Fig.~\ref{fig:ptools}). We use \textit{superquadrics} (Sec.~\ref{sec:superquadrics}) to model each part.

We break a p-tool vector $p$ into the following sections

\begin{align*}
\begin{split}
& p \phantom{aaaaa|} = [grasp,action,ga_{vec},a_{orient},m] \\
& grasp \phantom{|||} = [g_{a1},g_{a2},g_{a3},g_{\epsilon1},g_{\epsilon2},g_{Kx},g_{Ky},g_k,t] \\
& action \phantom{a} =  [a_{a1},a_{a2},a_{a3},a_{\epsilon1},a_{\epsilon2},a_{Kx},a_{Ky},a_k,t] \\
& ga_{vec} \phantom{aa} = [vc_x, vc_y, vc_z] \\
& a_{orient} \phantom{a} = [ao_{\phi},ao_{\theta},ao_{\psi}]
\end{split}
\end{align*}

where

\begin{itemize}
\item $grasp$ defines the scale ($g_{a1},g_{a2},g_{a3}$), shape ($g_{\epsilon1},g_{\epsilon2}$), tapering ($g_{Kx},g_{Ky}$), bending ($g_k$) and type ($g_t$) of the grasping part
\item $action$ defines the scale ($a_{a1},a_{a2},a_{a3}$), shape ($a_{\epsilon1},a_{\epsilon2}$) and tapering ($a_{Kx},a_{Ky}$), bending ($a_k$) and type ($a_t$) of the action part
\item $ga_{vec}$ defines a vector that goes from the centre of the grasping part to the centre of the action part ($vc_x, vc_y, vc_z$), the length of the vector determining the distance between the parts
\item $a_{orient}$ defines $ZYZ$ Euler angles for the action part general orientation (relative to its axes)
\item $m$ is the tool's mass
\end{itemize}

In \refabfig{ptools} we show various p-tools extracted from the ToolWeb dataset. The ladle (lower right corner) and mug (centre) show the use of bending and paraboloids. The chopstick (upper right corner) shows the use of tapering and also how a p-tool can be composed of only one superquadric (in this case, we repeat the superquadric twice in the p-tool vector.

The scales of the grasping and action part and the vector between the parts' centres is in metres; the mass is in kg. In order to be able to simulate using a p-tool we need to have a way of rendering it into a mesh model for the simulator. We are able to render each part using our sampling method for superquadrics \citep{AbelhaSQ2017}. We render the p-tool in its canonical representation, which is with the grasping part's centre at the origin and its orientation upwards (i.e. Euler angles set to $0$). From this we are able to put the action part at ($vc_x, vc_y, vc_z$) and rotate it according to ($ao_{\phi},ao_{\theta},ao_{\psi}$).


\subsubsection{P-tool Extraction}
\label{sec:ptool_extraction}

\begin{figure}[h]
	\centering	
	\includegraphics[width=8cm]{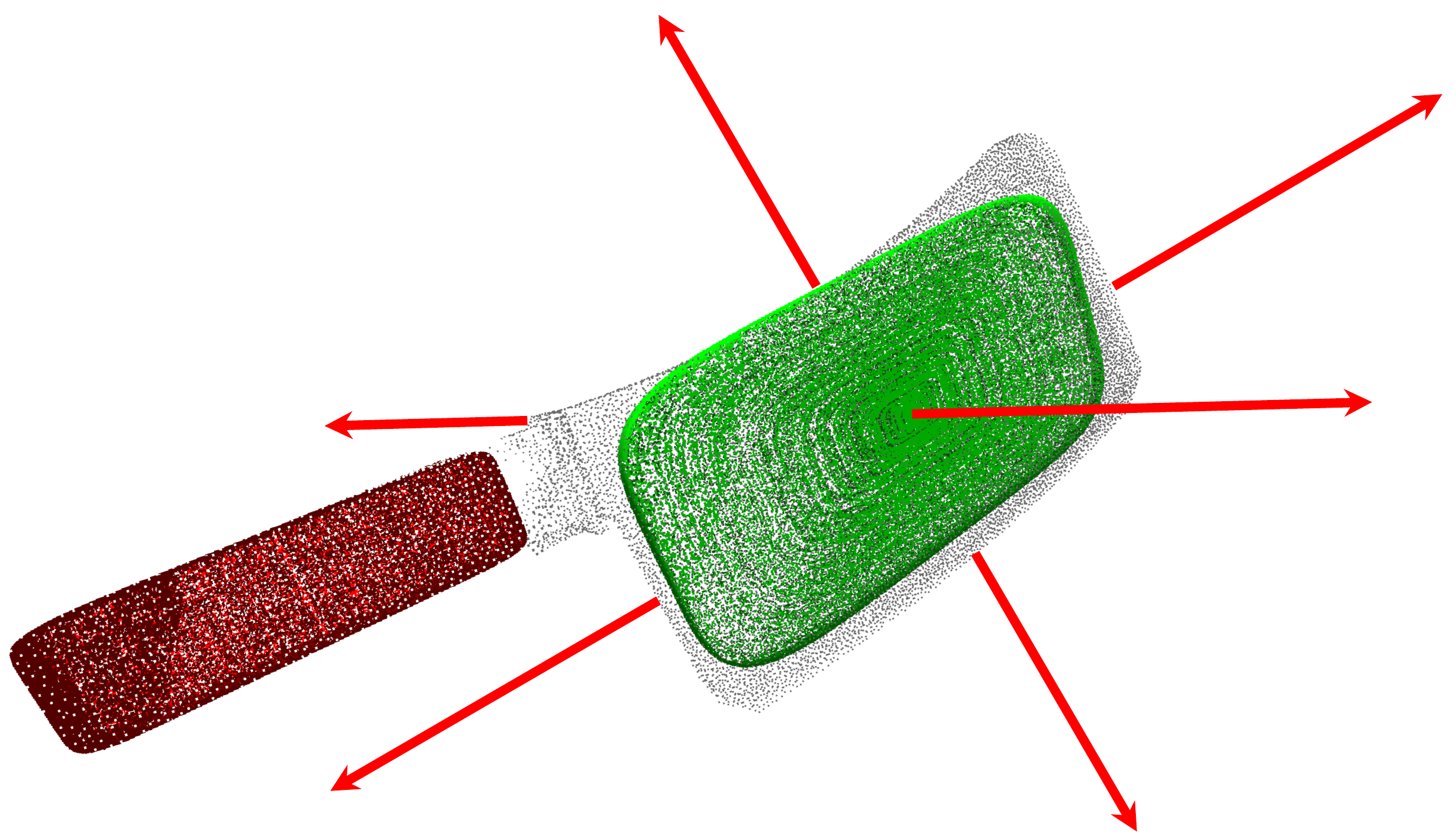}
	\caption{Alternative orientations for the action part (green) of a p-tool extracted from a Chinese knife (grasping in red). Each action part orientation leads to a different tool-use.}
	\label{fig:ptool_alt_fits}
\end{figure}

To generate our training p-tools we extract various p-tools form each segmented training point cloud. We are able to map a point cloud to a p-tool by first fitting superquadrics to its segments. The superquadric fitting is nondeterministic which leads to different fits in each run. Also, a particular fit of the action part determines its orientation and thus tool-use (see Sec.~\ref{sec:ptool_positioning}), but tools may be used in different ways. We create the new tool-uses by rotating the superquadric around each of its axes by $\pi/2$ and $-\pi/2$, with a total of 5 rotations plus original orientation: 6 orientations \refabfig{ptool_alt_fits}. We also change the scale parameters to adjust the superquadric. We do not create new rotation fits for grasping.

In Fig.~\ref{fig:ptool_alt_fits} we see six possible orientations for the action part; each of them implies a different way of using the Chinese knife for a task. For instance, if the task is cutting lasagne, it means that we would have the tool-uses: cutting with the (usual) long sharp edge; cutting with the front of the blade; cutting with the top of the blade; and cutting by pressing with either flat side of the blade (which doesn't work for cutting, but could be an effective way to crush garlic). There is also a fit which would use the back of the blade which is attached to the handle; we also include this.

The maximum number of p-tools obtained from a point cloud is $rn(n-1)$, where $r=6$ is the number of rotations and $n$ the number of segments. Not all rotations lead the superquadric to still have a good fit, therefore we accept only rotations that keep the fitting score to a maximum of $5\%$ more than the original score (the lower the fitting score the better). A point cloud with three segments can produce from $6$ to $216$ p-tools depending on how well each superquadric rotation fitted to the point cloud.

%
%


\subsubsection{P-tool Positioning}
\label{sec:ptool_positioning}

Each task requires a different relative positioning and orientation of the p-tool in the simulator. A p-tool contains all required parameters for positioning and rotation of a tool in the simulator for any task. To make it generic and possible to simulate for any given p-tool we devised a way of positioning any tool for any task given $12$ task-dependent parameters. For positioning we have the initial robot arm vector ($3$ parameters); and the absolute position of the task target contact point ($3$ parameters), e.g., nail's centre for hammering a nail; or dough edge for rolling dough. For rotation we have the task action vector ($3$ parameters) and task plane, defined by a normal vector to the plane ($3$ parameters).

We have as reference a `robot elbow' that is always positioned somewhere in the task simulation and represented by a small white sphere (visible in \refabfig{hammering_nail_sim} and \refabfig{scooping_grains_sim}). It is from this elbow that we define the joint that will be used in the task movement. The `robot arm' defines where the tool is going to be positioned relative to the elbow and is a vector that goes from the centre of the elbow to the centre of the grasping part. For all tasks we assume a robot arm length of $30$ cm.

For positioning we place the elbow relative to the task target contact point. Then we position the tool relative to the elbow by positioning its grasping part's centre at the end of the robot arm vector. For instance, in the case of hammering a nail, we want the tool's action part to land on the top of the nail after a $90$ degree rotation of the arm. This will depend on the size of the action part and its angle and distance to the grasping part. For each task we are able to define a simple function that positions the elbow taking into account the p-tool parameters.

For rotation we first align the ptool's action superquadric's $z$ vector with the task action vector and then we align the p-tools' $ga_{vec}$ with the task plane. For example, in the case of hammering a nail, we have the action vector as $[0,0,1]^T$ and the task plane normal vector as $[0,1,0]^T$.

\subsubsection{P-tool Inertial Parameters}
\label{sec:ptool_inertial}

In order to accurately simulate a p-tool in Gazebo we need to calculate its inertial parameters: MOI (moment of inertia) and centre of mass. We are able to get both the centre of mass, volume and MOI for a single superquadric in general position by using the derivations in \cite{JacklicSQBook}. Assuming homogeneous distribution of mass, we can calculate the centre of mass of the whole p-tool, composed by its superquadrics, and then calculate the MOI relative to the p-tool's centre of mass.\footnote{Gazebo has a method to calculate the MOI of an object, but it is too coarse being based on using a bounding box and calculating the MOI for that box.}

\subsection{Task Simulation}
\label{sec:task_simulation}
We have five tasks: \textit{hammering nail}; \textit{lifting pancake}; \textit{rolling dough}; \textit{cutting lasagne}; and \textit{scooping grains}. For each task we create a simulation version of it (in Gazebo\footnote{http://gazebosim.org/}) in which we define the actions associated with performing the task. For every task the simulator outputs a real number scoring the result (e.g. how much the nail went down for hammering a nail); we run the simulation three times for each tool and take the median as the final result. We convert this real number to the $1-4$ range in which the dataset ToolWeb was labelled \refabsecp{toolweb_dataset}.

\subsubsection{Calibration}
\label{sec:task_calibration}

We calibrate the simulator to make it a good proxy for the real-world task using the complete ToolWeb dataset \refabsecp{toolweb_dataset}. For this we can change the task simulation parameters for the task, e.g., the friction of the nail's joints for hammering nail task \refabsecp{hammering_nail_simulation}. We can also change the $3$ thresholds for the function we use to categorise the real-valued output for the simulation into the $1-4$ range \refabsecp{toolweb_dataset}. We alter the task simulation parameters and the function thresholds until we achieve at least $90\%$ accuracy on ToolWeb. 

\subsubsection{Hammering nail}
\label{sec:hammering_nail_simulation}

\begin{figure}[h!]		
	\centering
	\includegraphics[width=4cm]{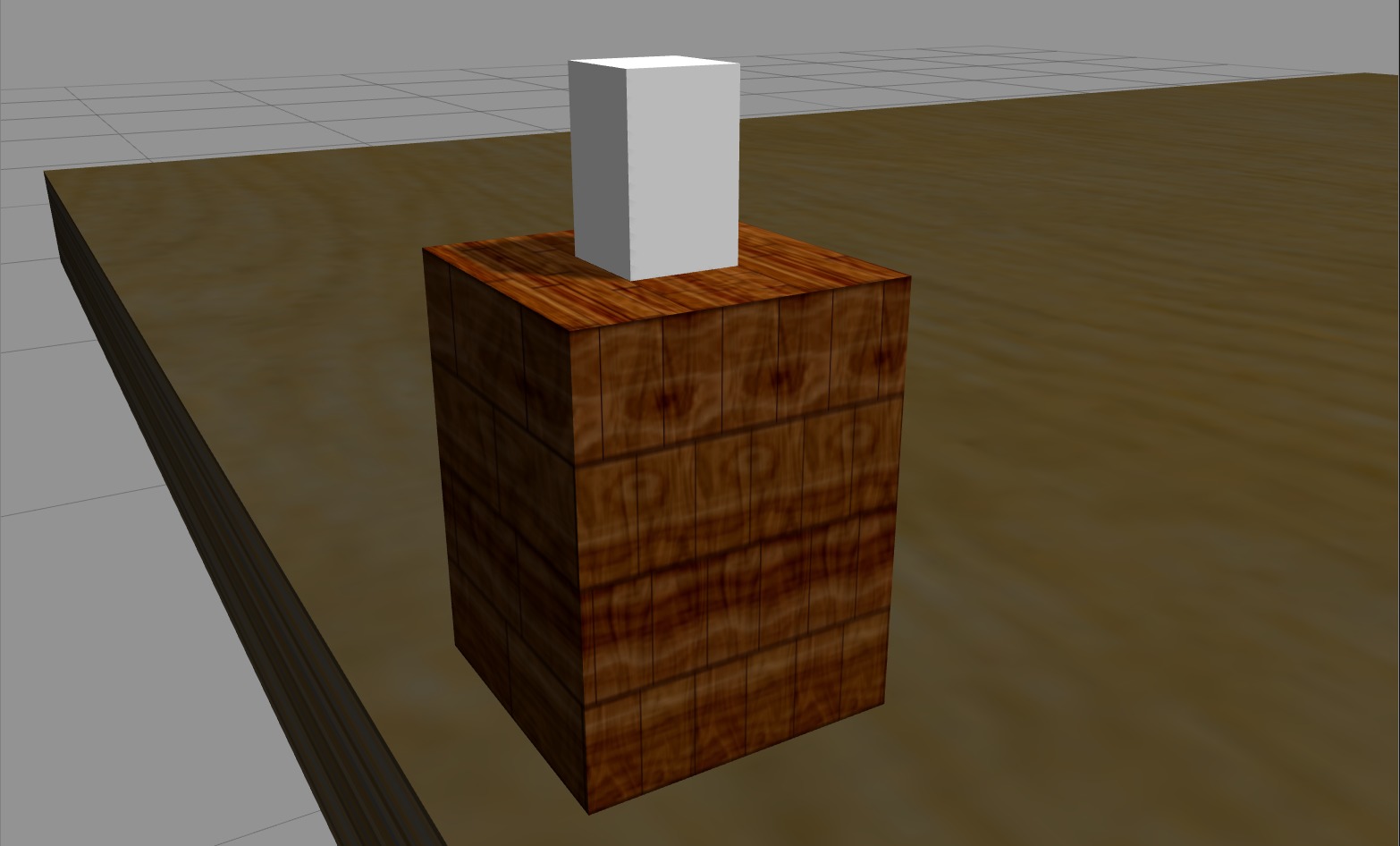}
	\includegraphics[width=4cm]{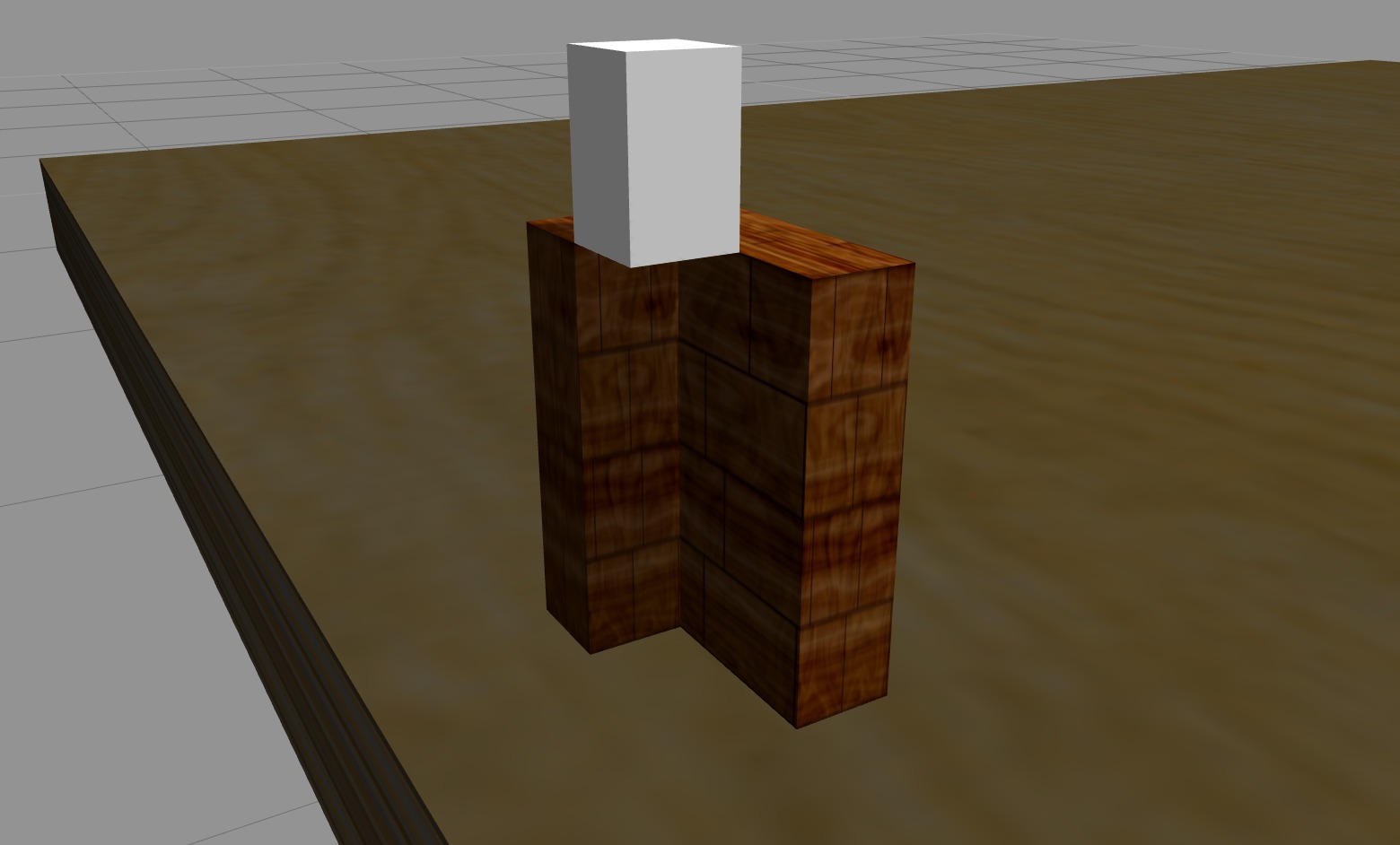}
	\caption{Nail model for the hammering nail task. Model is made of four wooden blocks and one nail block. On the left is the full model; on the right, we remove two wooden blocks to show how the nail can `penetrate' the wood by sliding through the hole formed by the four blocks.}
	\label{fig:nail_model}
\end{figure}

We constructed the nail model (Fig.~\ref{fig:nail_model}) by building a box model where four boxes are fit so as to leave a hole in the middle, where another box, the nail, can penetrate. The nail box also has prismatic joints with each of the four blocks. The friction of the joints simulate how hard it is to push the nail `through' the wood. The two hyper-parameters for calibration are the nail mass and the friction at the joints.

\begin{figure}[h!]		
	\centering
	\includegraphics[width=4cm]{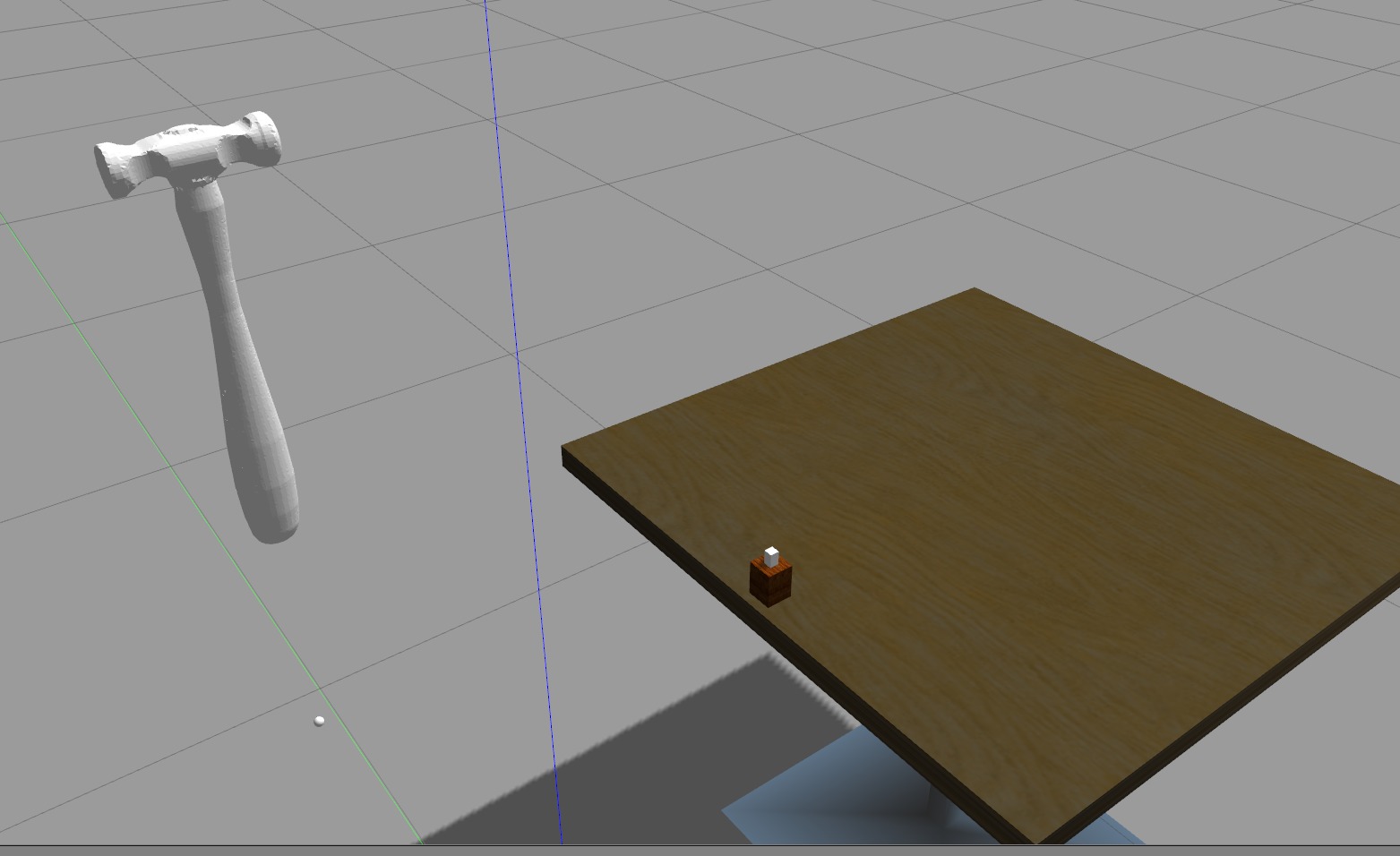}
	\includegraphics[width=4cm]{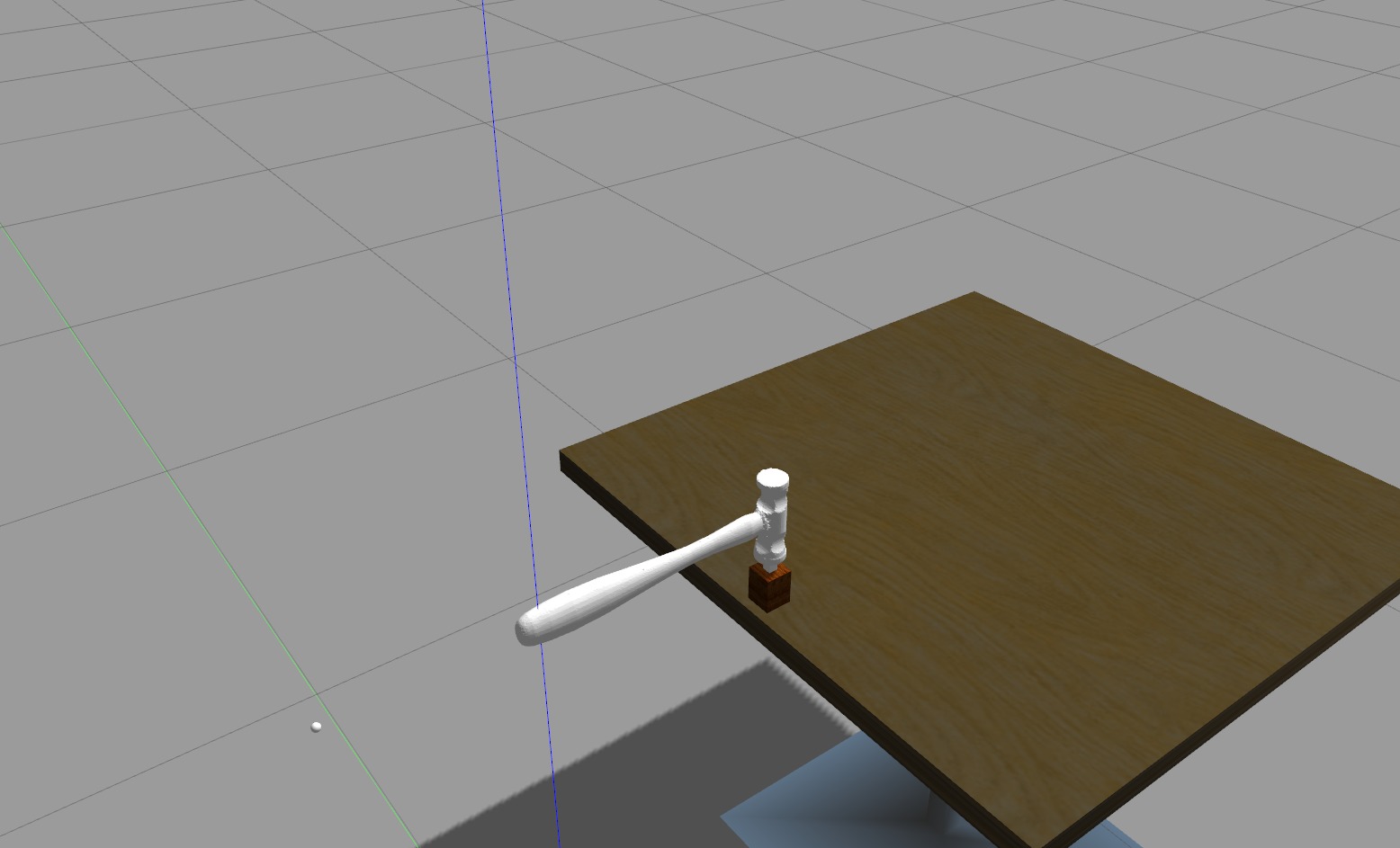}
	\caption{Task simulation for hammering a nail performed with a hammer from our ToolWeb dataset.}
	\label{fig:hammering_nail_sim}
\end{figure}

For the hammering nail simulation we perform one rotation with fixed torque (Fig.~\ref{fig:hammering_nail_sim}). After the tool has collided with the nail, we measure how much the nail goes down and use this as the output for the simulation.

\subsubsection{Lifting pancake}
\label{sec:lifting_pancake_simulation}

\begin{figure}[h!]		
	\centering
	\includegraphics[width=4cm]{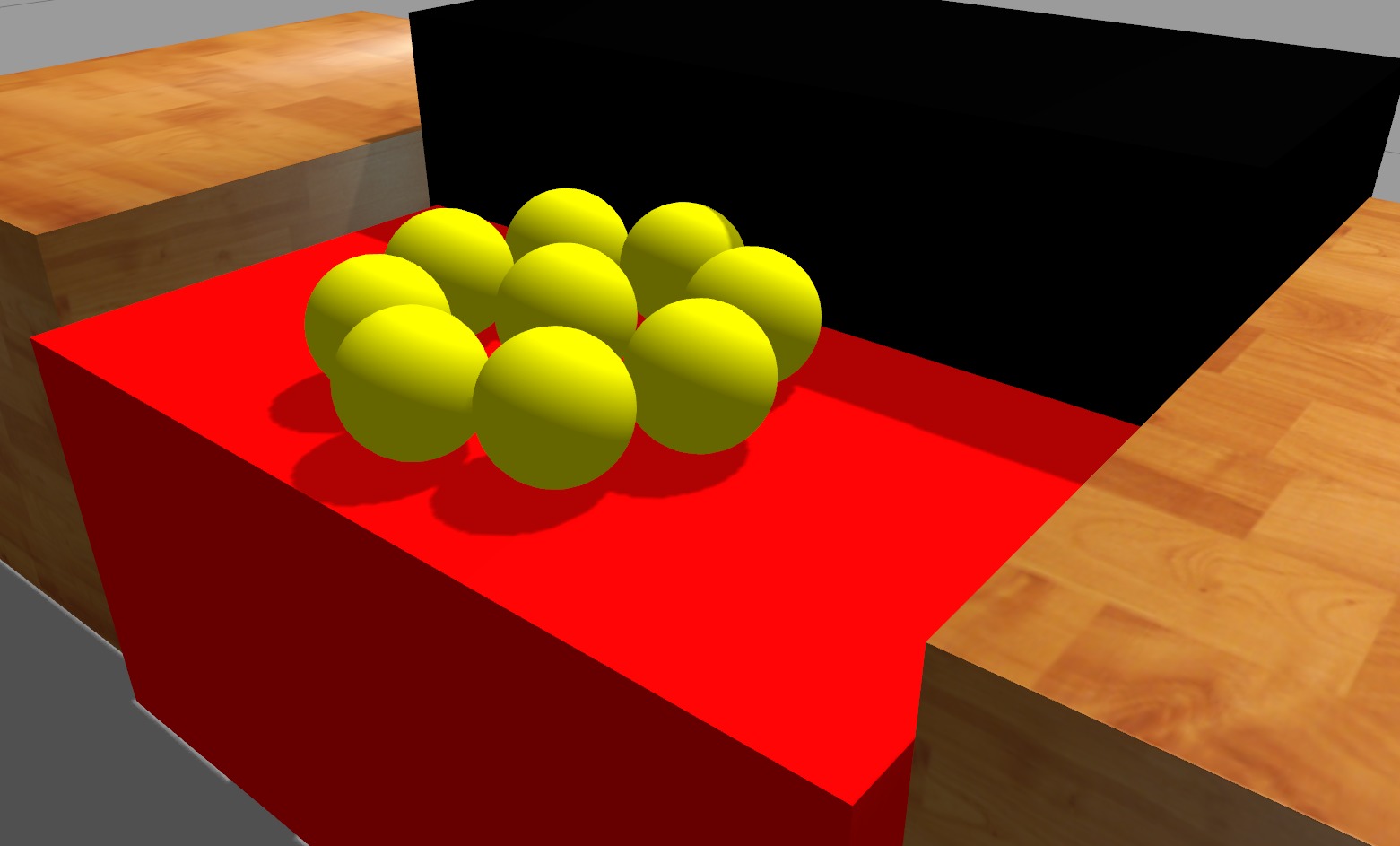}
	\caption{Pancake model composed of $9$ spheres connected through invisible flexible joints.}
	\label{fig:pancake_model}
\end{figure}

We constructed the pancake model (Fig.~\ref{fig:pancake_model}) with a circle of $8$ spheres, with $1$ central one, and they are all connected through invisible flexible joints. Before calibration, we experimented with different physics parameters of the spheres until we achieved the best behaviour for the pancake in order to simulate the viscosity and `hardness'. The two hyper-parameters for calibration are the pancake overall mass (divided among the spheres) and the friction at the spheres' joints.

\begin{figure}[h!]	
	\centering
	\includegraphics[width=4cm]{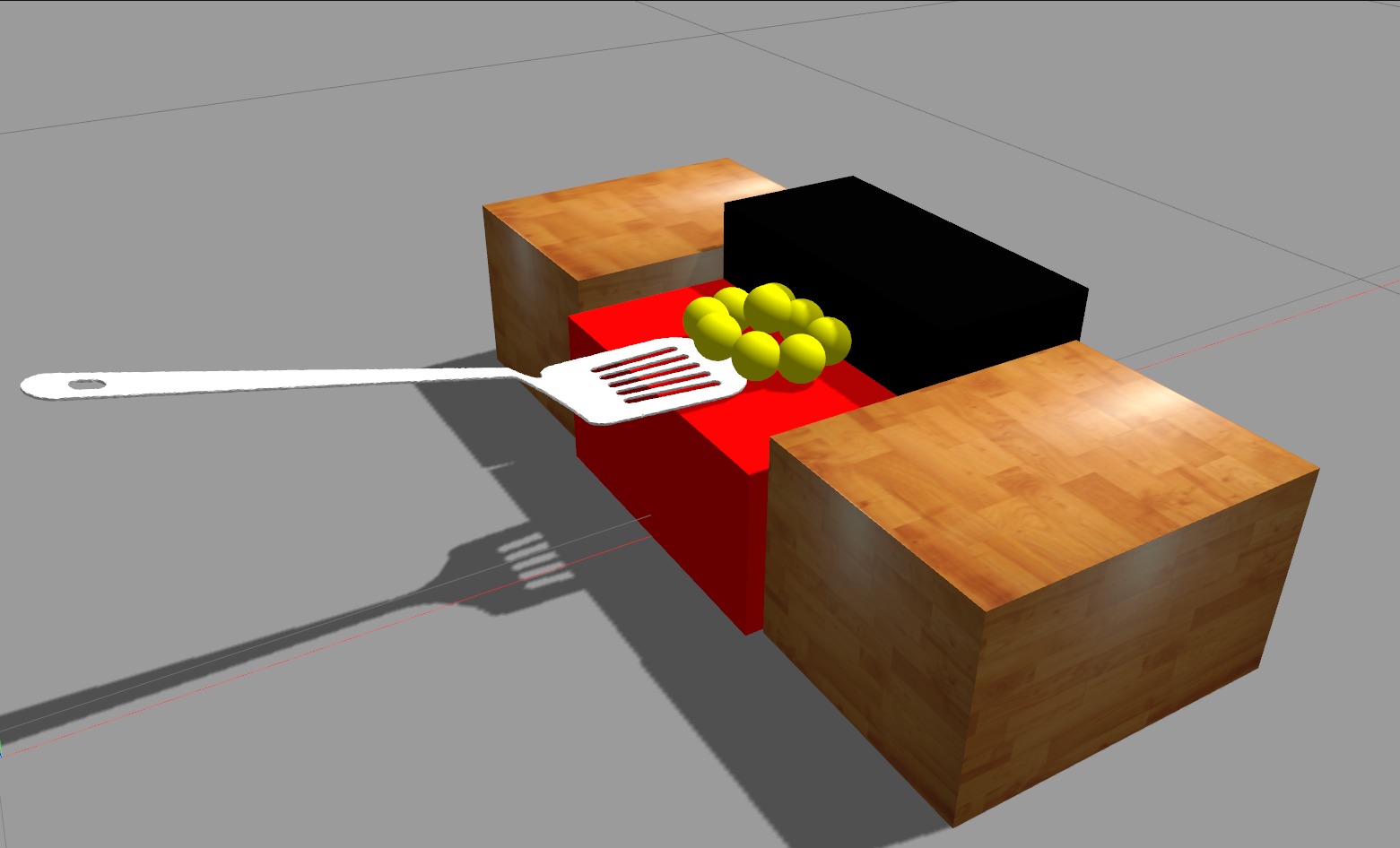}
	\includegraphics[width=4cm]{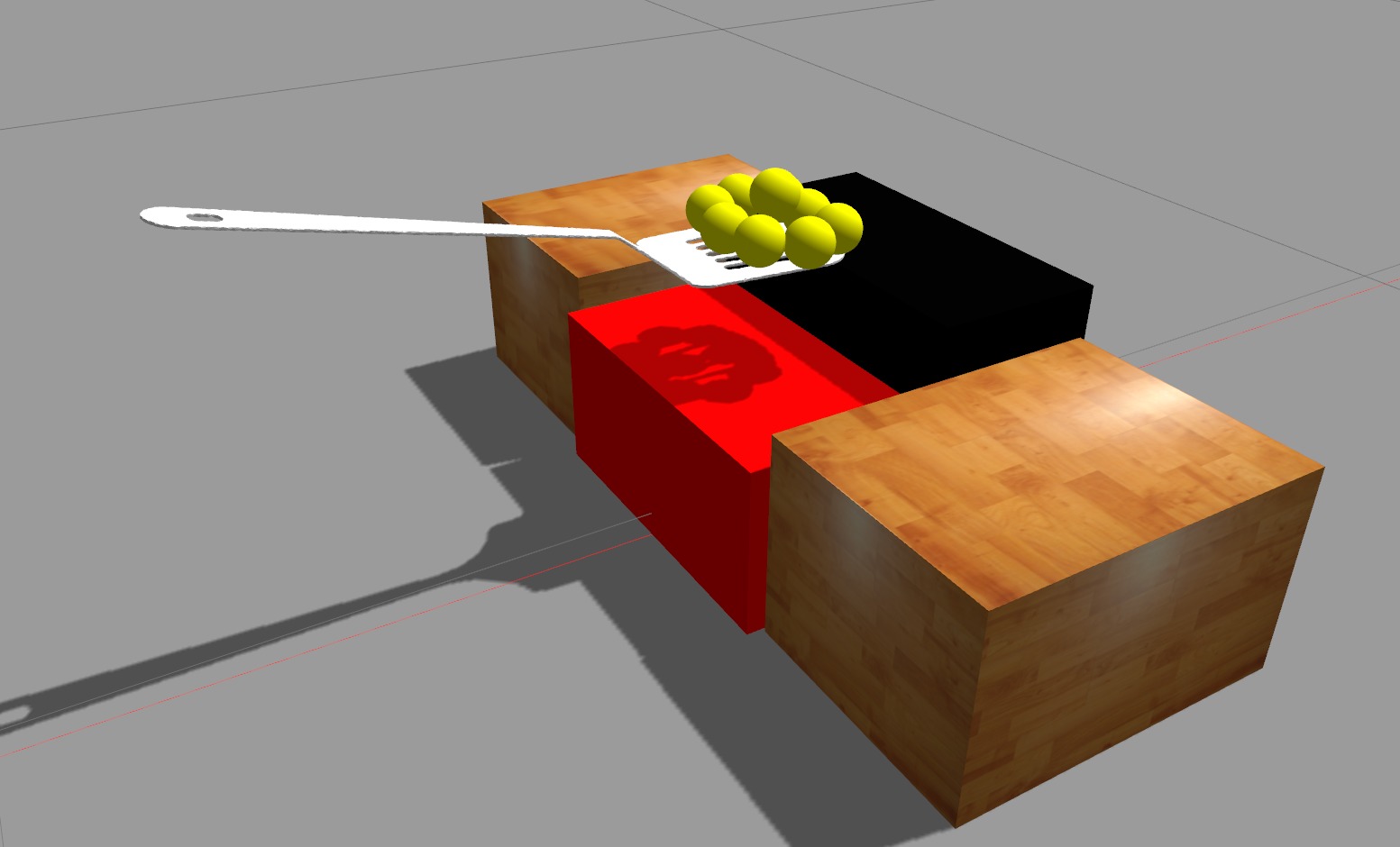}
	\caption{Task simulation for lifting pancake performed with a mesh spatula from our ToolWeb dataset.}
	\label{fig:lifting_pancake_sim}
\end{figure}

For lifting pancake simulation we perform a straight motion towards the pancake and then a lifting motion after the tool touches the black box (Fig.~\ref{fig:lifting_pancake_sim}). We measure the tool's effectiveness by measuring the disfigurement of the pancake and also if its centre was lifted with the tool. We calculate how close the pancake's spheres' centres are to a plane, and also if the pancake's central sphere is above and close to the tool's action part's centre, in order to determine the output for the simulation.

\subsubsection{Rolling dough}
\label{sec:rolling_dough_simulation}

\begin{figure}[h!]	
	\centering
	\includegraphics[width=4cm]{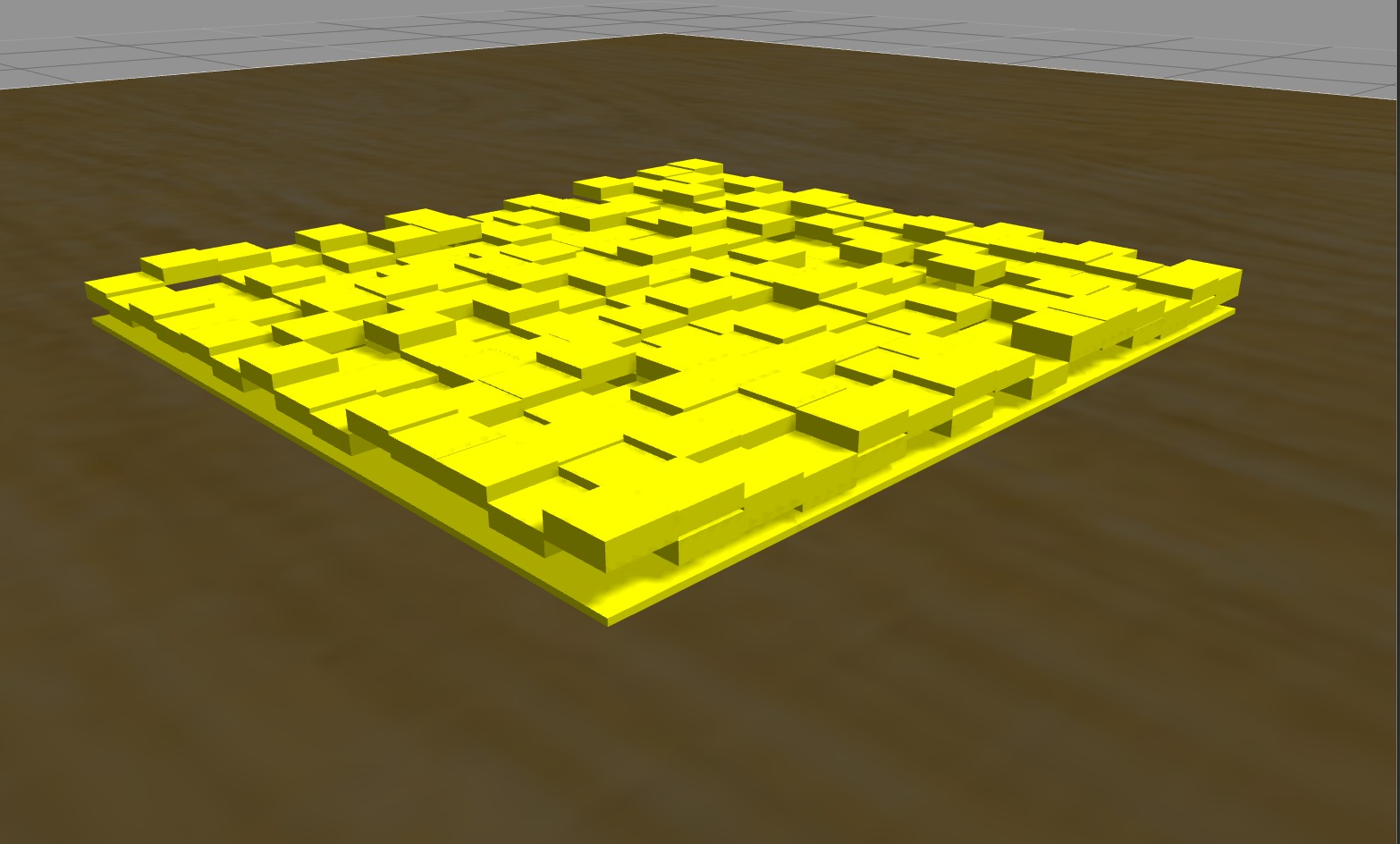}
	\caption{Dough model composed of a flat box with various blocks that start at a random height and are connected to it through joints.}
	\label{fig:dough_model}
\end{figure}

We built the dough model composed of a very heavy flat box and 225 (15x15) blocks starting at random heights (Fig.~\ref{fig:dough_model}) on top of the flat box. Each block has a prismatic joint connecting it to the flat box. The damping and friction on each joint simulate the `hardness' of the dough to resist flattening. The two hyper-parameters for calibration are the damping and the friction.

\begin{figure}[h!]	
	\centering
	\includegraphics[width=4cm]{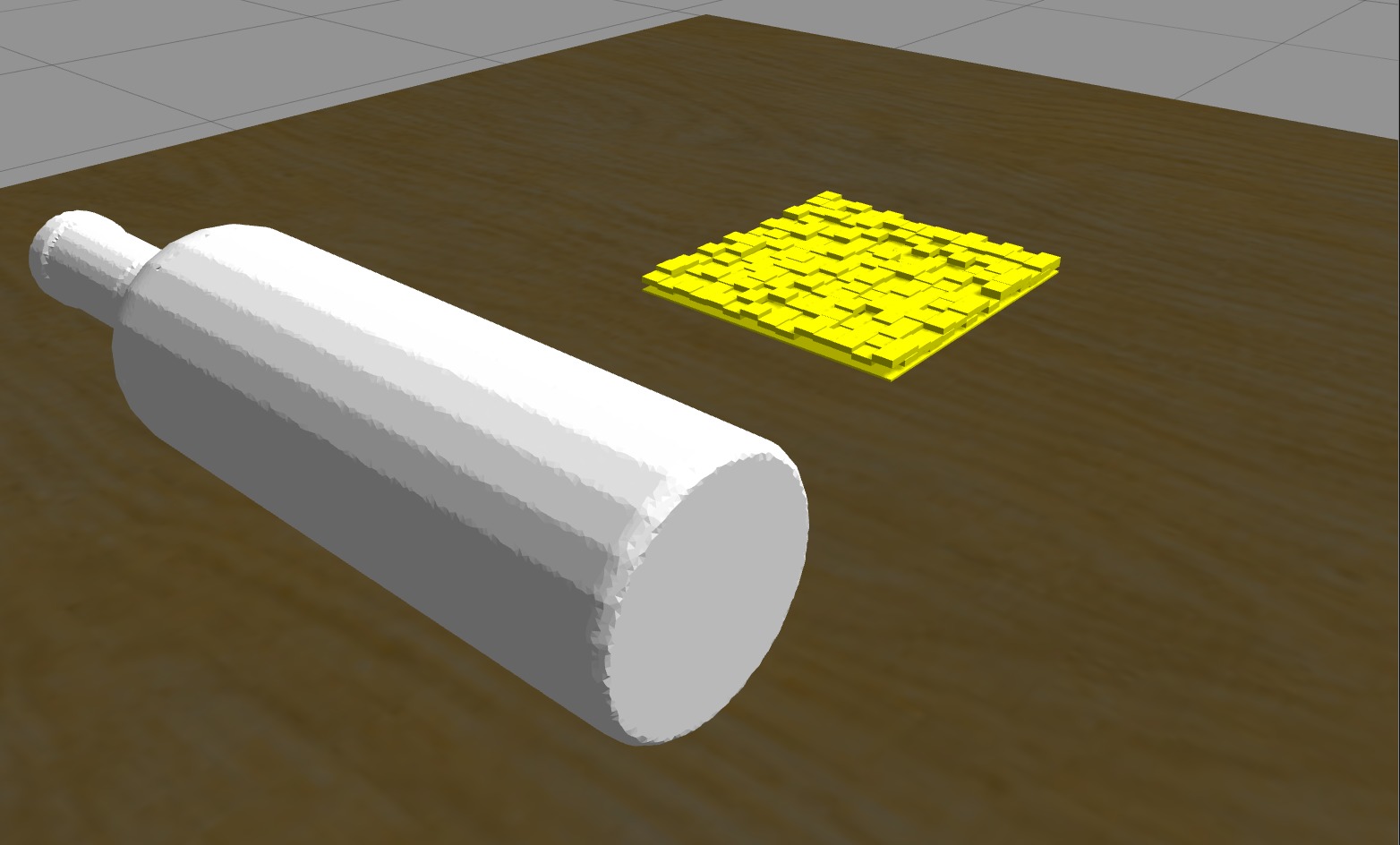}
	\includegraphics[width=4cm]{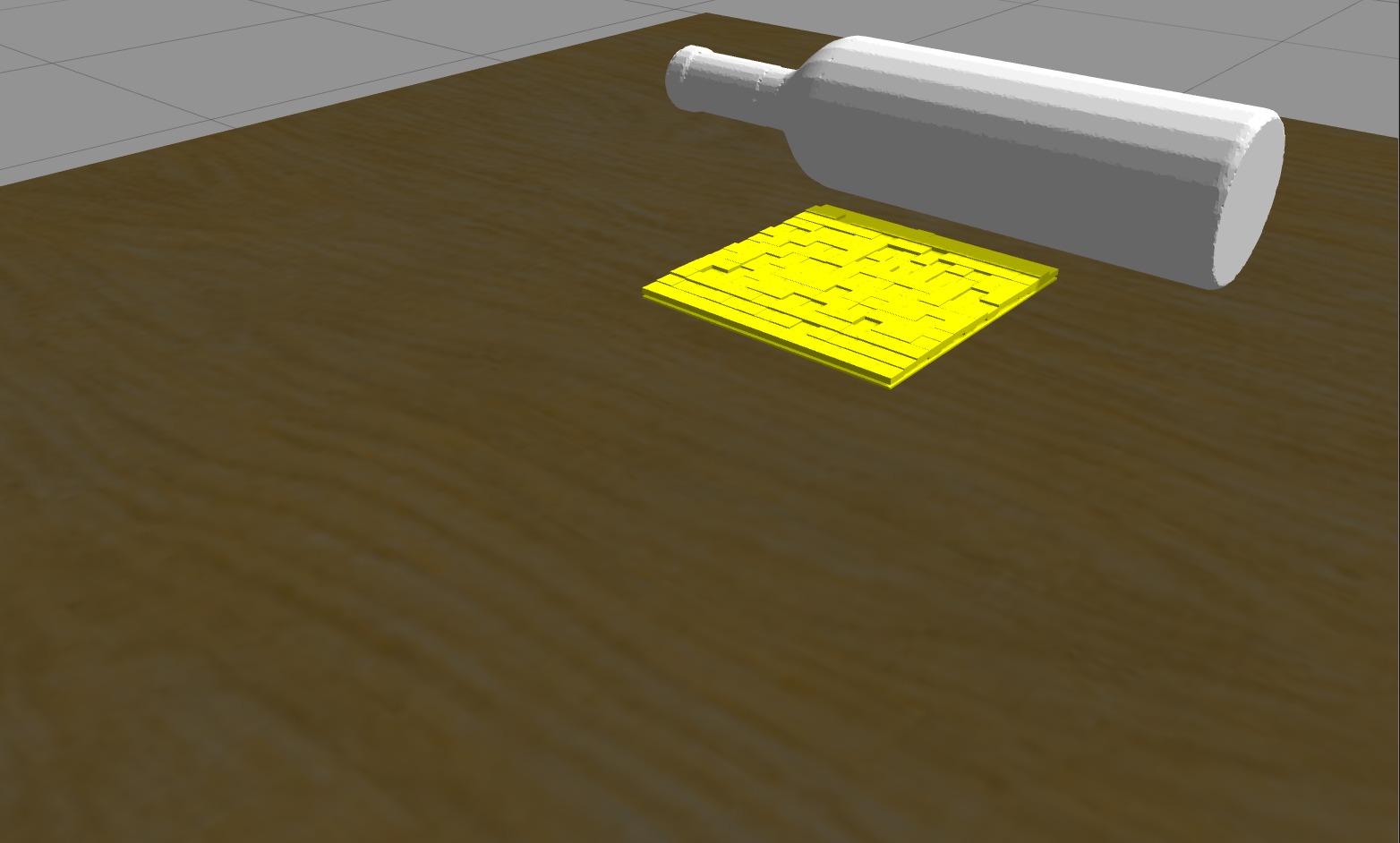}
	\caption{Task simulation for rolling dough performed with a wine bottle from our ToolWeb dataset.}
	\label{fig:rolling_dough_sim}
\end{figure}

For rolling dough simulation we perform a straight motion towards the dough combined with angular velocity (to `roll' the tool) and downward force (Fig.~\ref{fig:rolling_dough_sim}). We calculate the variance of the dough's boxes' centre position in order to determine the output for the simulation.

\subsubsection{Cutting lasagne}
\label{sec:cutting_lasagne_simulation}

\begin{figure}[h!]	
	\centering
	\includegraphics[width=4cm]{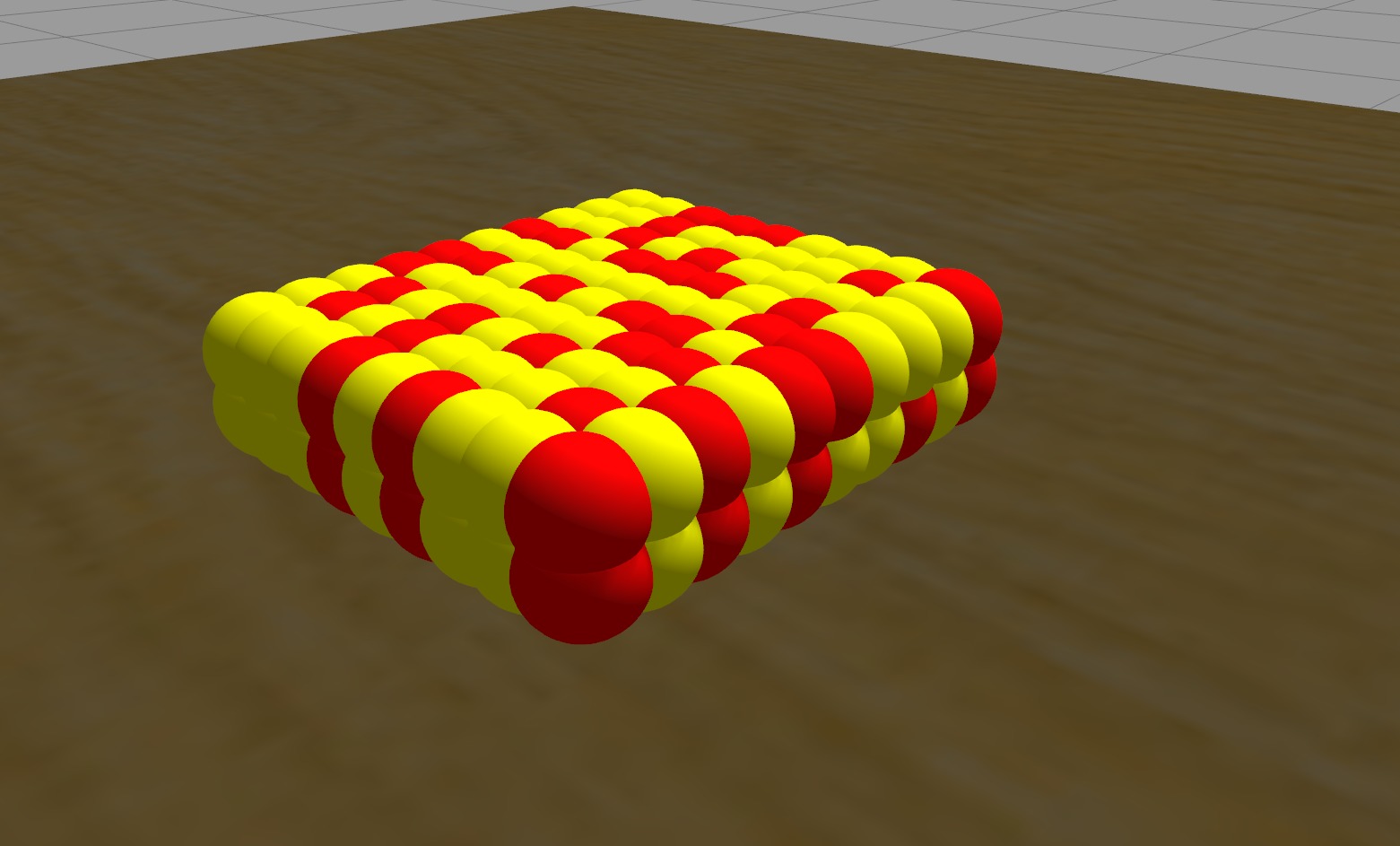}
	\caption{Lasagne model composed of $10\times10\times2$ spheres connected through joints.}
	\label{fig:lasagne_model}
\end{figure}

We built the lasagne model composed of $10\times10\times2$ spheres (Fig.~\ref{fig:lasagne_model}) connected through invisible flexible joints. Similarly to the pancake, before calibration we  experimented with different physics parameters of the spheres until we achieved the best behaviour also to simulate the viscosity and `hardness' of the lasagne. The two hyper-parameters for calibration are the damping and the friction of the spheres' joints.

\begin{figure}[h!]	
	\centering
	\includegraphics[width=4cm]{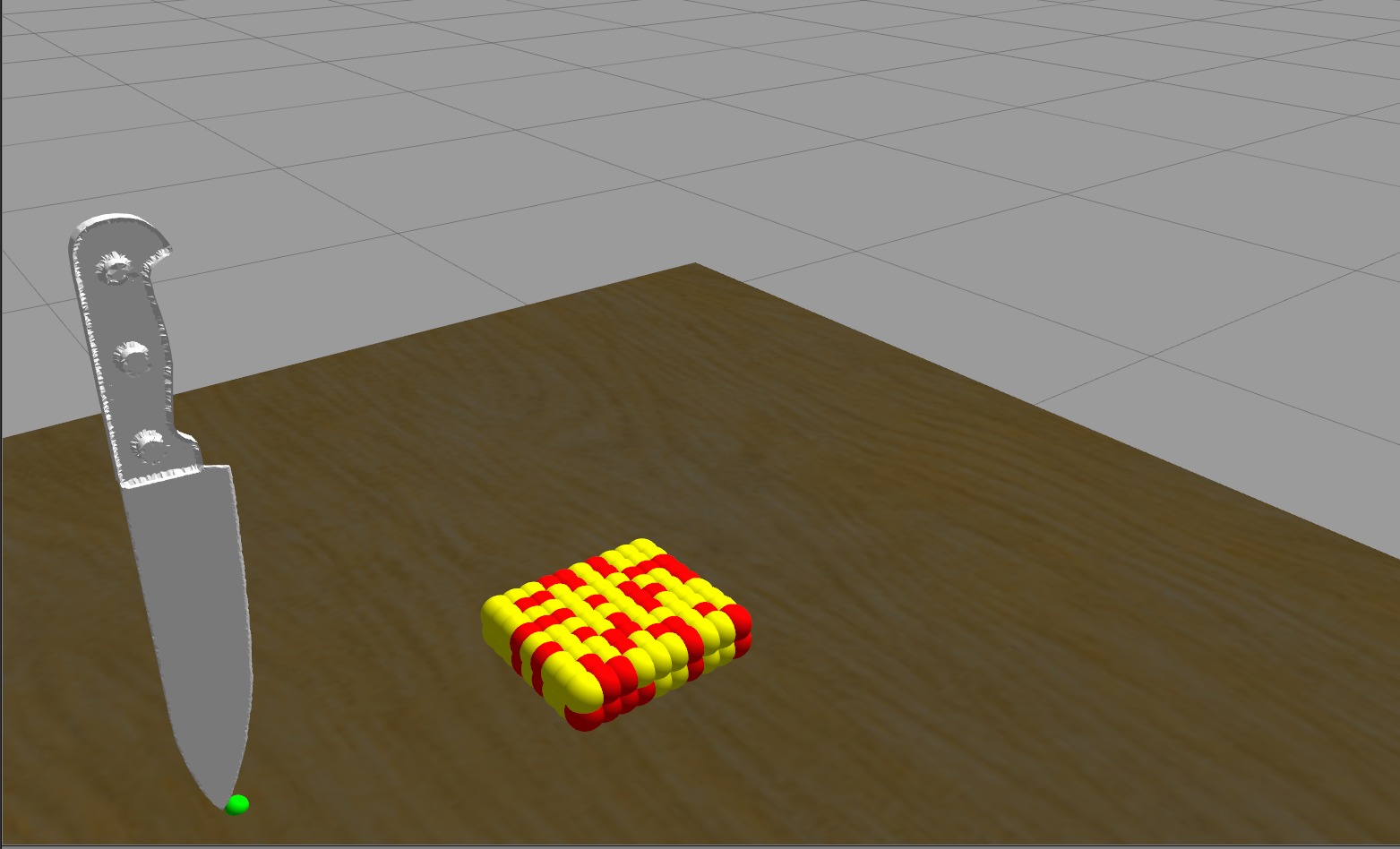}
	\includegraphics[width=4cm]{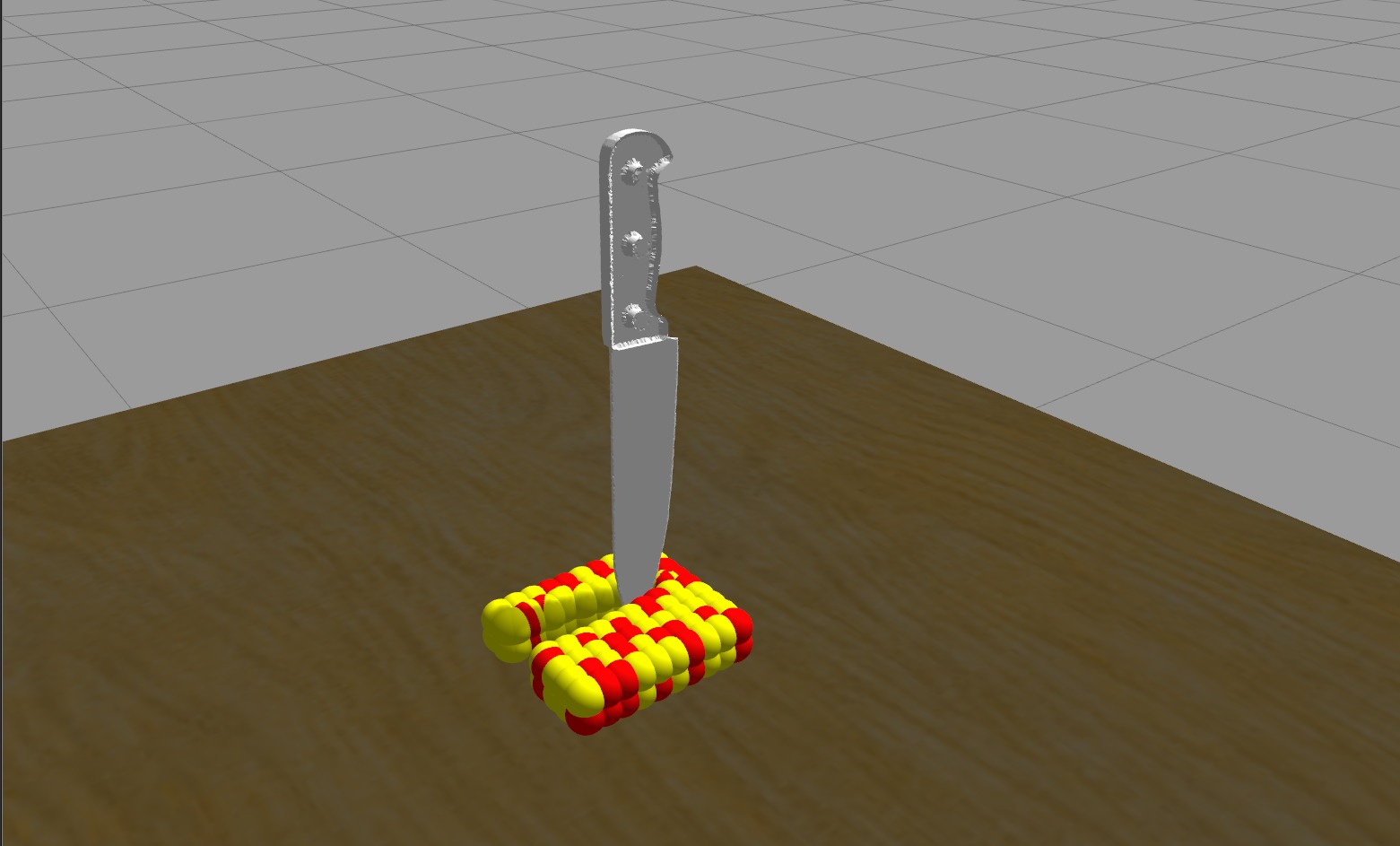}
	\caption{Task simulation for cutting lasagne performed with a kitchen knife from our ToolWeb dataset.}
	\label{fig:cutting_lasagne_sim}
\end{figure}

For cutting lasagne simulation we perform a straight motion towards and across the lasagne \refabfig{lasagne_model}. We calculate how much each sphere has been displaced from its initial position and use the inverse of the median displacement as the output for the simulation.

\subsubsection{Scooping grains}
\label{sec:scooping_grains_simulation}

\begin{figure}[h!]	
	\centering
	\includegraphics[width=4cm]{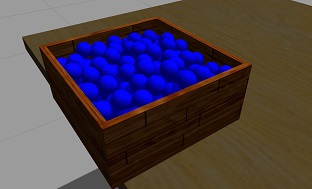}
	\caption{Grains model composed of $100$ independent spheres in a box.}
	\label{fig:grains_model}
\end{figure}

We built the grains model by putting $100$ independent and unconnected spheres in a box \refabfig{grains_model}. The hyper-parameter for calibration is the mass for each sphere.

\begin{figure}[h!]	
	\centering
	\includegraphics[width=4cm]{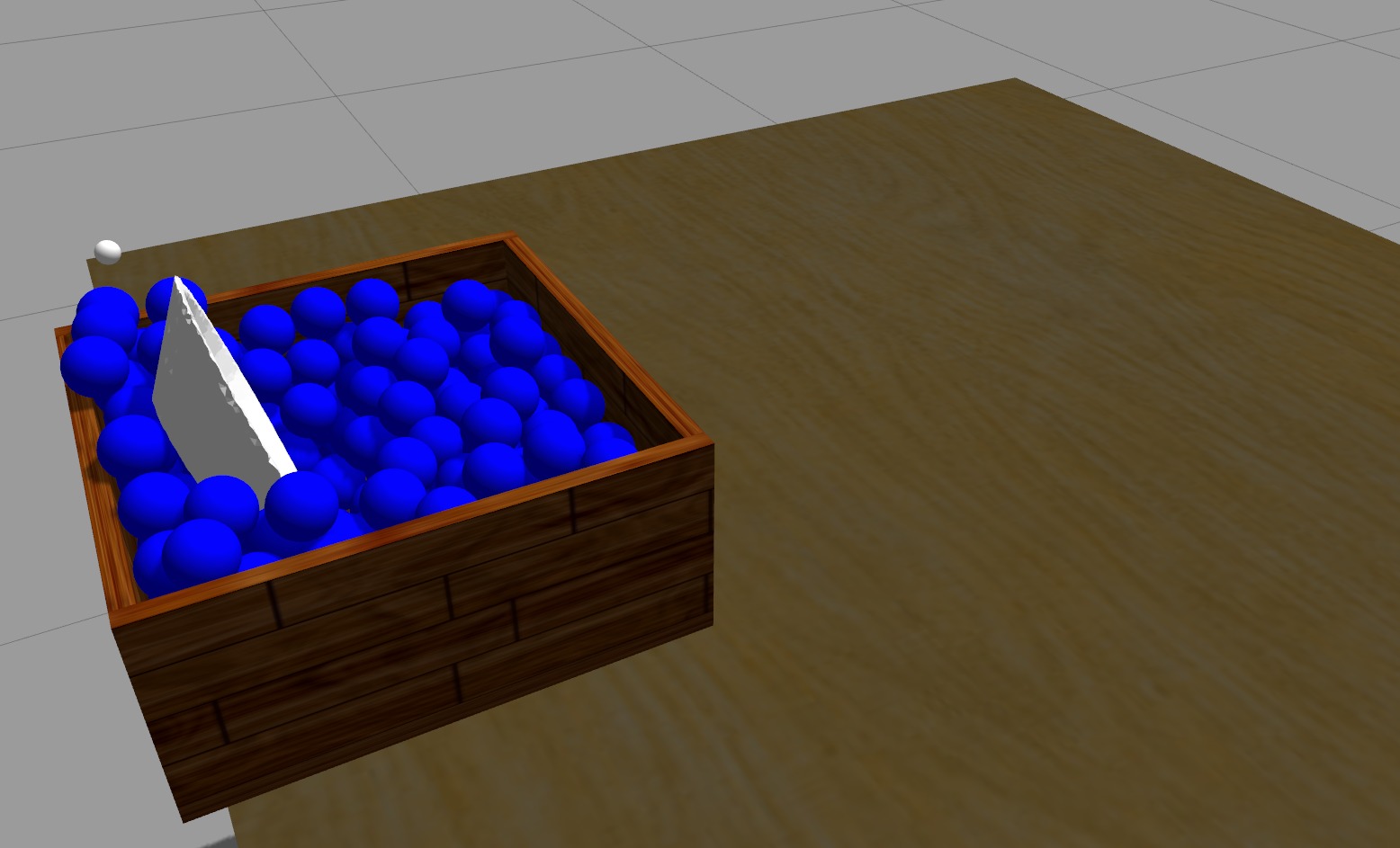}
	\includegraphics[width=4cm]{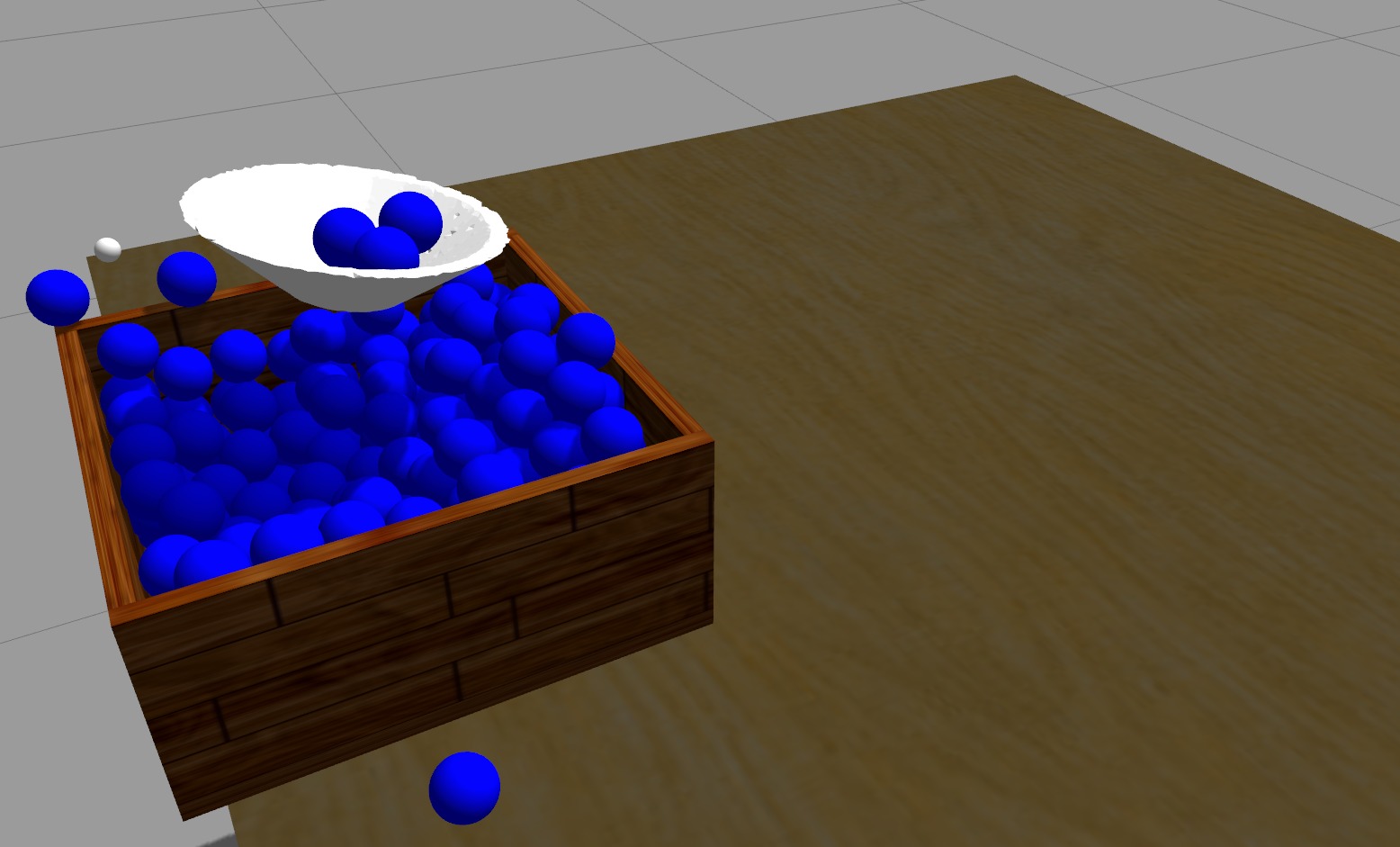}
	\caption{Task simulation for scooping grains performed with a bowl from our ToolWeb dataset}
	\label{fig:scooping_grains_sim}
\end{figure}

For scooping grains simulation we perform one rotation in order to bring the tool down and another inverse rotation to bring it up  (Fig.~\ref{fig:scooping_grains_sim}). Both rotations are done with fixed torque. We calculate the number of scooping grains by counting the grains that are above the tool's action part's centre discounting grains that are too far away (to avoid grains that might be too high due to strong forces). The number of scooped grains is the output for the simulation.

\subsection{Task Function Learning}
\label{sec:task_function_learning}

For each task we learn a function that maps a p-tool vector to an affordance score by performing a Gaussian Process Regression (GPR) \citep{RasmussenGPBook} on a set of $2185$ p-tools labelled by task simulations. The GPR's training set is the $2185$ p-tools with added inertia (diagonal of the moment of inertia tensor) and centre of mass ($25+3+3=31$ dimensions) and the labels are the scores coming from the simulator for each p-tool. We train the GP using Matlab's Statistics and Machine Learning implementation \footnote{https://uk.mathworks.com/help/stats/gaussian-process-regression-models.html}.

The training data is balanced in order to improve training and have a uniform distribution over affordance scores: the categorised affordance scores are obtained through a threshold function applied on real-valued simulation outputs \refabsecp{task_calibration}. We generate more p-tools by running an interpolation sampling on the $2185$ p-tools. This interpolation works by sampling $100$ p-tools in between every pair of p-tools \textit{iff} the Euclidean distance between them is smaller than the mean distance between each p-tool and its closest neighbour. The closest neighbour to a p-tool is  the one with smaller Euclidean distance to the p-tool (if there is more than one closest neighbour than a random one is chosen).

After sampling the interpolated p-tools we filter them so they fall in a range of pre-specified valid parameters for tools. These new p-tools are then assessed by the trained GPR model trained on the $2185$, throwing away excess p-tools until we achieve $5000$ p-tools with an approximate uniform distribution over affordance scores (i.e. no affordance score has more than 5\% more p-tools than another one). These balanced $5000$ p-tools are then simulated. After simulation we learn the final task function through GPR on the balanced $5000$ p-tools and their real-valued simulation outputs.

We use the well-known squared exponential kernel with Automatic Relevance Determination (ARD) to train for every task. The ARD kernel allows us, after training, to get the \textit{length-scale} for each dimension of the training data \citep{RasmussenGPBook}. These length-scales informally represent how much you need to walk in that dimension before the value of the function changes significantly. That is, the length-scales tell you how much each dimension contributes to the function value. A large length-scale means that the dimension has small influence over the function values. It is important to note, however, that the length-scale will be in the same scale as the corresponding values for that dimension of the data.

Taking advantage of the ARD we run iterations of the GPR, each time calculating an `importance value' for each dimension and eliminating those that have a low value. To calculate a dimension's importance we divide the its range of values by the length-scale to get a value that is proportional to the contribution of that dimension while also accounting for the length-scale being in the same ``units'' as the dimension. We stop training when all length-scales have their importance values above $1$. 

\subsection{Assessing a New Tool}

\subsubsection{Overview}
\label{sec:assessing_new_tool_overview}

We want to assess how good a new previously unseen tool is for a task. A point cloud will have many ways of being abstracted and we are interested in the ones in which the p-tool has a good fit to the point cloud and is also good for the task (has a good task function score).

\subsubsection{Projection}
\label{sec:projection}
We call a projection the assignment of a p-tool to a point cloud under a given task. The idea of projection in general is to find a good match between the bottom-up realities of the raw data and the top-down pressure from some task. When we map a raw point cloud to a p-tool there is a trade-off between representing the local shapes in the point cloud well, and having a good score in the task function.  When assessing a new tool we do not consider its segments, but provide a comparison with using them in our ablation studies \refabsecp{exp_ablation}.

For projecting we have five steps:

\begin{enumerate}
\item Planting spheres centered on randomly chosen seed points on the point cloud, with ten radii chosen randomly between $0.001$ and $0.1$ meters. Form this we get a set of \textit{seeded point clouds}, i.e., each point cloud that has its points inside each sphere.
\item Superquadric fitting for each seed \refabsecp{superquadric_fitting}
\item Extracting p-tools from the fitted superquadrics, considering possible rotations \refabsecp{ptool_extraction}; and getting a fit score for the p-tool as the sum of the fitting score for both superquadrics
\item Evaluating the task function for every p-tool \refabsecp{task_function_learning}
\item Performing a weighted voting on fit and task score and picking the best p-tool
\end{enumerate}

In step $2$ the fitted superquadrics to the seed point clouds can give rise to a different number of p-tools. For instance, $20$ seeds will lead to fitting $20$ superquadrics and getting from $380$ to $2280$ different p-tools \refabsecp{ptool_extraction}. Step $5$, weighted voting, is done after normalising both the fitting scores and the task function values for each p-tool. The values for both weights are normally set at $1$, but we allow the weights to change in order to perform ablation studies \refabsecp{exp_ablation}.

The best candidate p-tool will then contain the tool-use parameters and can be used to assess the tool for grasping and also for positioning and orientation for the task. This p-tool can be readily aligned with the point cloud from which it was abstracted, therefore the parameters are ready to use. The grasping parameters are the superquadric representing the grasping region of the tool. It is important to note that the p-tool representation is intuitive for humans so our approach could also be used as a module to fit in a larger framework. The p-tool's geometric intuitive interpretation makes it easy to build frameworks on top of it that reason over aspects of the tool such as scale, shape, inertia and mass.
\section{Experimental Evaluation}

In this section we first present our test sets \refabsecp{exp_test_sets} and then three different experimental evaluations, First \refabsecp{exp_comp_scw}, a comparison between our system and the closest one in the literature \citep{SchoelerTCDS2016}, which we here refer to as \textit{ScW}. Second \refabsecp{exp_ablation}, we include ablation studies, measuring performance when modifying aspects of the system. Finally \refabsecp{exp_run_times}, we measure running times for various parts of the system; only those that are implemented by us, excluding for instance, learning the task function through the Matlab GPR implementation. 

We can run our system with different weights for the fitting function score and the task function score \refabsecp{assessing_new_tool_overview}. The usual setting is to have both weights set to 1 (i.e. a balance between abstraction and assessment). In the ablation studies we change the weight for the task function to 0 in order to test the role of the task function in getting results. Whenever we set both weights to 1 we will refer to this as an equal balance between fitting score and task function \refabtabp{comparison_toolartec1}.

\subsection{Test Sets}
\label{sec:exp_test_sets}

Our system is evaluated on three datasets: ToolArtec, ToolArtecSmall and ToolKinect \footnote{Datasets available at: github.com/pauloabelha/ToolWeb /ToolArtec and /ToolKinect}. 
The first dataset is ToolArtec and it consists of 50 full-view point clouds of real household objects scanned with the high-end Artec Eva 3D scanner. For each tool we obtained a full-view by fixing it on the ground, moving the scanner all around the object and using Artec Eva's real-time fusion, with default settings for smooth fusion and hole filling. A small number of objects were problematic, such as the large frying pans. These required two scans from different view points which were later fused using Artec Eva's software to perform registration. All of this scanning required human judgement (e.g. during real-time fusion to observe which side of the tool was still missing and to scan that side), however existing work in robotics has shown that a robot can autonomously obtain such complete point clouds by picking up an object and rotating it, then putting it down and regrasping from another point, in order to complete parts that had been obscured by the robot's hand previously \cite{KraininIJRR2011}.

In order to obtain an objective assessment of the affordances of the objects they were tried out on tasks and a resulting effect measured, for four tasks: For hammering a small nail was hammered in softwood, and the depth to which it was driven was measured; For lifting pancake a proxy was created with child's play dough making a circular pancake of 160 mm diameter and 2 mm thickness; For rolling dough a sphere of child's play dough was rolled, and the time taken to achieve a satisfactory result (if possible at all) was measured; for cutting lasagne a cylinder of child's play dough was cut, and the deformation was measured. For these tasks the measured values tended to cluster, and so we categorised the results with four categories, defined by thresholds. For example in hammering the tool was scored as 4 (best) if the nail was driven $>=$ 10 mm; score 3 if $>$ 5 mm; score 2 if $>$ 1 mm; score 1 (worst) if
$<=$ 1mm. For 14 of the 50 tools labels were extrapolated from very similar tools that had been tried. For the final scooping task the labels were given by human judgement.

\begin{figure}[ht]	
	\centering	
	\includegraphics[width=8cm]{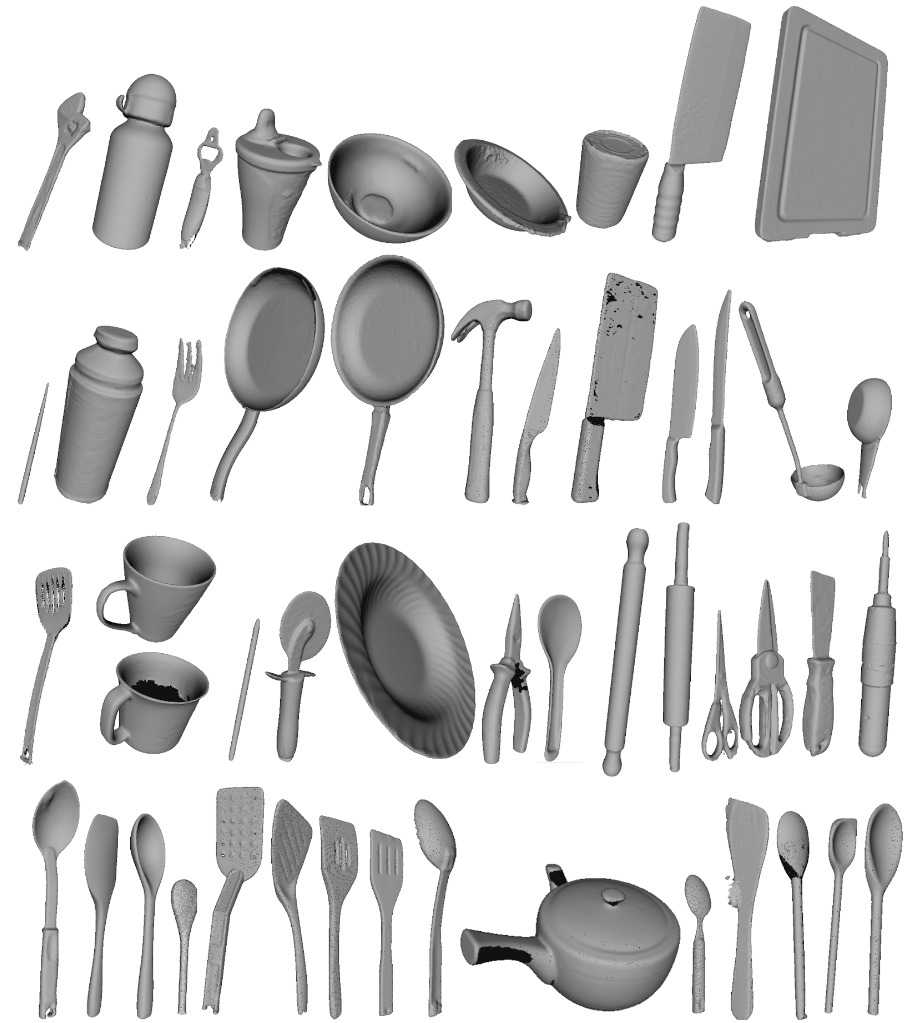}
	\caption{The ToolArtec dataset composed of 50 full-view point clouds gathered with the Artec Eva 3D scanner. The dataset has mass and affordance score groundtruth for 5 everyday tasks.}
	\label{fig:ToolArtec}
\end{figure}

The second dataset is ToolKinect and it is a subset of 13 point clouds taken from the 50 tools from ToolArtec, but scanned with the Kinect 2 scanner. This is important to compare our results given high and low-quality scanners. The main reason why this dataset is so small is that it is very hard to reliably obtain good quality point clouds from shiny or glossy object using Kinect 2 \citep{AbelhaICRA2016}. 

\begin{figure}[ht]	
	\centering	
	\includegraphics[width=8cm]{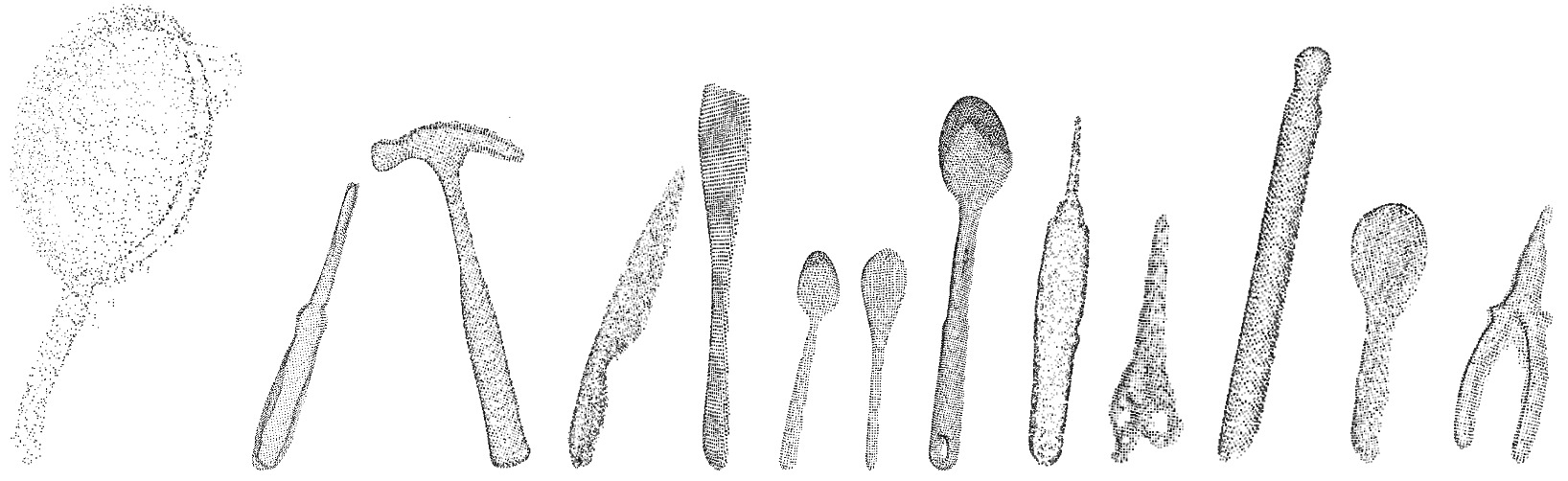}
	\caption{The ToolKinect dataset composed of 13 point clouds gathered with the Kinect v2 scanner. The dataset has mass and affordance score groundtruth for 5 everyday tasks.}
	\label{fig:ToolKinect}
\end{figure}

The third dataset is ToolArtecSmall and it is just a subset of ToolArtec point clouds from the same tools present in ToolKinect. ToolArtecSmall allows us to directly compare running our method in high and low quality scanners. Labels for ToolKinect and ToolArtecSmall are inherited from ToolArtec. More details on both datasets can be found in cite{AbelhaICRA2016}.

It is important to note that ToolKinect and ToolArtecSmall are too small to provide a thorough evaluation and they are not representative of a reasonable range of kitchen tools. With the improvement of sensor technology and reduction of scanner cost, we expect much larger datasets from household tools (that can contain shiny or glossy objects).

\subsection{Comparison with ScW}
\label{sec:exp_comp_scw}

We compare our work with ScW \citep{SchoelerTCDS2016} by training ScW with 15 tools (of affordance score 4) in four tasks \footnote{The number 15 was recommended as a minimum by Markus Schoeler}. We could not train ScW for the scooping grains task because the system failed when dealing with many of the affordance score 4 scooping tools (e.g bowls and plate). It would not be fair then to compare our system with a trained ScW that missed many of the good tools for scooping. ScW is trained in a partially different set of tools than used in the comparison in \cite{AbelhaIROS2017} because now we have a different ToolWeb dataset with 116 point clouds. We achieve $69\%$ overall accuracy on four different everyday tasks compared to ScW that achieves $32\%$ on the same four tasks.

There are two metrics used in the comparisons: \textit{Metric 1} and accuracy. Accuracy is just the proportion of tools for which the system guessed the precise affordance score; we calculate accuracy for each of the four affordance scores. Metric 1 is a squared difference between system estimate and ground truth:

\begin{align*}
	m_1 = \frac{(c-1)^2-\frac{1}{n}\sum_{i=1}^{n} |\textbf{s}_i-\textbf{g}_i|^2}{(c-1)^2}
\end{align*}

where $c>1$ is the number of possible discrete (affordance) scores (in our case $4$), $n$ is the number of scores (size of $\textbf{s}$ and $\textbf{g}$), $\textbf{s}$ is the vector containing the system's scores, $\textbf{g}$ is the vector containing the ground truth scores. Metric 1 is always between $0$ and $1$ and quadratically penalises the distance between the scoring vector and the ground truth.

Additionally to comparing with ScW we also include a baseline score for each task composed of random ratings. For each task, each tool is given a random affordance score (1-4); we run this $10^5$ times, each time computing Metric 1 using the randomly generated scores and the groundtruth for each task.  
From this we get $10^5$ Metric 1 values for each task that roughly follow a normal distribution. We take the ($\mu$) and standard deviation ($\sigma$) of the values and, assuming a normal distribution, calculate what the $p$-value ($p<0.05$) for each task. That is, the Metric 1 value above which there is only a $5\%$ or less chance of achieving that by randomly scoring each tool for the task. Table \ref{tab:comparison_toolartec1} shows the $\mu$, $\sigma$ and $p$-values for each task and averaged over all tasks for ToolArtec. Table \ref{tab:overall_all_datasets} shows these values averaged over all tasks for each dataset.

\begin{table*}[ht]
\caption{Metric 1 comparison of our system with ScW \cite{SchoelerTCDS2016} on ToolArtec using 20 seeds and an equal balance between fitting score and task function.}
\centering
\begin{tabular}{ l | c | c | c | c | c }
	\hline\hline
&\multicolumn{2}{c|}{Metric 1}&\multicolumn{3}{c}{Random ratings}\\ 

&ours&ScW&$\mu$ 	&$\sigma$	&$p<0.05$	 	\\ \hline

ToolArtec   		&0.91&0.83&0.72&0.04&0.77
 \\ \hline

Hammering   		&0.94&0.80&0.77&0.03&0.82
\\ \hline

Lifting   		&0.88&0.84&0.73&0.05&0.81
\\ \hline

Rolling   		&0.93&0.86&0.66&0.04&0.73
\\ \hline

Cutting   		&0.86&0.83&0.70&0.05&0.78
\\ \hline

Scooping   		&0.92&-&0.70&0.05&0.77
\\ \hline
\end{tabular}
\label{tab:comparison_toolartec1}
\end{table*}

\begin{table*}[ht]
\caption{Accuracy comparison of our system with ScW \cite{SchoelerTCDS2016} on ToolArtec using 20 seeds and an equal balance between fitting score and task function. Cells in red show where our system has lower score than ScW for the given task.}
\centering
\begin{tabular}{ l | c | c | c | c | c | c | c | c }
	\hline\hline
&\multicolumn{2}{c|}{Accuracy}&\multicolumn{2}{c|}{Accuracy}&\multicolumn{2}{c|}{Accuracy}&\multicolumn{2}{c}{Accuracy}\\ 
&\multicolumn{2}{c|}{Aff. score 1}&\multicolumn{2}{c|}{Aff. score 2}&\multicolumn{2}{c|}{Aff. score 3}&\multicolumn{2}{c}{Aff. score 4}\\ 

&ours&ScW&ours&ScW&ours&ScW&ours&ScW	 	\\ \hline

ToolArtec   		
&0.75&0.47&0.62&0.13&0.74&0.43&0.73&0.22 \\ \hline

Hammering   		
&0.63&0.53&0.55&0.00&0.60&0.40&0.75&0.25 \\ \hline

Lifting   		
&\cellcolor{red!25}0.58&0.65&0.67&0.20&\cellcolor{red!25}0.64&0.78
&0.63&0.13 \\ \hline

Rolling   		
&0.81&0.34&0.55&0.33&0.67&0.00&1.00&0.00 \\ \hline

Cutting   		
&1.00&0.38&0.50&0.00&0.82&0.56&0.57&0.50 \\ \hline

Scooping   		
&0.75&-&0.85&-&1.00&-&0.71&- \\ \hline

\end{tabular}
\label{tab:comparison_toolartec2}
\end{table*}

\begin{table}[h]
\caption{Overall Metric 1 on all test sets using 20 seeds and an equal balance between fitting score and task function.}
\centering
\begin{tabular}{ l | c | c | c | c | c }
	\hline\hline
	&\multicolumn{2}{c|}{Metric 1}&\multicolumn{3}{c}{Random ratings} \\
				&our & ScW 		&$\mu$ 		&$\sigma$	&$p<0.05$	\\ \hline
	ToolArtec   &$0.91$&$0.83$	&$0.72$ 		&$0.04$ 		&$0.77$		\\ \hline
	SmallArtec  &$0.90$&$0.86$	&$0.79$ 		&$0.05$ 		&$0.85$		\\ \hline
	ToolKinect	&$0.86$&$0.77$ 	&$0.75$		&$0.03$		&$0.80$   	\\ \hline
\end{tabular}
\label{tab:overall_all_datasets}
\end{table}		

In Table~\ref{tab:overall_all_datasets} we present overall results for the three dataset. We can see that our approach obtains better overall scores in all datasets and that both our approach and ScW are able to be above the $p$-values for all datasets. Also, we obtain better results for all datasets as compared with \cite{AbelhaIROS2017}, which can be partially explained from the fact that we start with 116 tools in ToolWeb instead of 70.

Tables~\ref{tab:comparison_toolartec1} and \ref{tab:comparison_toolartec2} are the main result tables where we present the results for ToolArtec, including the results per task (Table~\ref{tab:comparison_toolartec1}) and per affordance category for each task (Table~\ref{tab:comparison_toolartec2}). We highlight results where we did worse than ScW by painting the corresponding cells in red.

We achieve better results than ScW in all tasks, excluding affordance categories 1 and 3 for the lifting pancake task. Our approach is worst at identifying affordance category 2, which is also the worst category for ScW. On the other hand, the accuracy of our approach is balanced between categories 1, 3 and 4. ScW had some problems with certain categories achieving 0 accuracy (e.g. category 4 in the rolling dough task). 
ScW only trains on category 4 (best tools); a larger set that had more such tools could be expected to improve the ScW score, especially improving its ability to discriminate category 4 from lower categories, but it will likely still struggle to make judgements for intermediate categories.

\subsection{Ablation}
\label{sec:exp_ablation}

In order to evaluate the importance of different aspects of our approach we perform  two ablation studies: trying our approach using and not using the task function to select the best candidate (Table~\ref{tab:w_wout_task_func}); and trying our approach with four different numbers of seeds: 5, 10, 15 and 20 (Table~\ref{tab:per_n_seeds}). We highlight results where we did worse than ScW by painting the corresponding cells in red.

\begin{table*}[ht]
\caption{Metric 1 on ToolArtec using 20 seeds and with an equal balance between fitting score and task function (\textit{Balanced fitting and task function} column) versus weight 0 for the task function (\textit{Only fitting score}). Column \textit{All tasks} shows the average for all tasks. We also present the random ratings values for each task and the average for all tasks. Cells in red show where our system has lower score than ScW for the given task.}
\centering
\begin{tabular}{ l | c | c | c | c | c  }
	\hline\hline
	&\multicolumn{2}{c}{selection of p-tool} 
	&\multicolumn{3}{c}{Random ratings}						\\
		&&Balanced fitting 	&&& 							\\ 
		&Only fitting score&and task function
		&$\mu$&$\sigma$&$p<0.05$ 							\\ \hline
	All tasks &0.85&0.91&0.72&0.04&0.77						\\ \hline
	Hammering &0.89&0.94&0.77&0.03&0.82						\\ \hline
	Lifting   &\cellcolor{red!25}0.83&0.88&0.73&0.05&0.81	\\ \hline
	Rolling   &\cellcolor{red!25}0.84&0.93&0.66&0.04&0.73	\\ \hline
	Cutting   &0.84&0.86&0.70&0.05&0.78						\\ \hline
	Scooping  &0.88&0.92&0.70&0.05&0.77						\\ \hline
\end{tabular}
\label{tab:w_wout_task_func}
\end{table*}	

In Table~\ref{tab:w_wout_task_func} we can see that without using the task function our system is worse than ScW in two tasks: lifting pancake and rolling dough. Lifting pancake is the task with the worst score for us and rolling dough has the largest improvement when using task function for candidate selection. In Table~\ref{tab:per_n_seeds} we see that our approach requires the use of 20 seeds in order to be better than ScW in every task; with 15 seeds we are only worse than ScW in rolling dough. Finally, also in Table~\ref{tab:per_n_seeds} we see that our approach improves by using more seeds (exploring more the space of p-tools), 
Although with smaller numbers the figures are more susceptible to variation (i.e. the seeds may get lucky sometimes), and need many trials to determine the average.

\begin{table*}[!ht]
\caption{Metric 1 on ToolArtec  when varying number of seeds and with an equal balance between fitting score and task function. Column \textit{All tasks} shows the average for all tasks. Cells in red show where our system has lower score than ScW for the given task.}
\centering
\begin{tabular}{ l | c | c | c | c | c | c | c }
	\hline\hline
	&\multicolumn{4}{c}{Number of seeds} 
	&\multicolumn{3}{c}{Random ratings}						\\
		&5&10&15&20	 					
		&$\mu$&$\sigma$&$p<0.05$ 							\\ \hline
	All tasks &0.84&	0.85&0.88&0.91&0.72&0.04&0.77	\\ \hline
	Hammering &0.89&0.92&0.93&0.94&0.77&0.03&0.82	\\ \hline
	Lifting   &\cellcolor{red!25}0.83&0.86&0.87&0.88&0.73&0.05&0.81	\\ \hline
	Rolling   &\cellcolor{red!25}0.82&\cellcolor{red!25}0.82&\cellcolor{red!25}0.84&0.88&0.66&0.04&0.73 \\ \hline
	Cutting   &\cellcolor{red!25}0.82&\cellcolor{red!25}0.81&0.84&0.86&0.70&0.05&0.78		\\ \hline
	Scooping  &\cellcolor{red!25}0.84&0.83&0.90&0.92&0.70&0.05&0.77		\\ \hline
\end{tabular}
\label{tab:per_n_seeds}
\end{table*}	


\subsection{Running Times}
\label{sec:exp_run_times}

All running times reported in this paper are obtained with the same desktop computer: Intel(R) Core(TM) i5-3470 CPU @ 3.20GHz. In order to fit nicely into columns, all result tables with task names use an abbreviation with just the task verb (e.g. hammering nail becomes hammering).

\subsubsection{Superquadric Fitting}
\label{sec:exp_sq_fitting}
For superquadric fitting \refabsecp{superquadric_fitting} the average over $100$ runs is $1.83\pm0.10$ seconds for a point cloud with a $1000$ points.

\subsubsection{Simulation}

The simulation time for each task is different. Averaging over $10$ different tools: hammering nail takes about $3$ seconds; lifting pancake, about $10$ seconds; rolling dough about $6$ seconds; cutting lasagne about $5$ seconds; and scooping grains about $10$ seconds. Lifting pancake and scooping grains are slower because the tool moves slowly when lifting the pancake and the grains model is quite heavy for a normal desktop without using GPU acceleration for the physics engine.

\subsubsection{Projection}

Running projection \refabsecp{projection} on the entire ToolArtec dataset takes around $0.25$ hours for 5 seeds; $0.51$ hours for 10 seeds; $2.12$ hours for 15 seeds; and $5.36$ hours for 20 seeds. Please recall that we are planting spheres with 10 different radii for each seed, so 20 seeds leads to 200 fittings, and then each possible pair is considered for grasping place and action part, in six orientations each.

\begin{figure}[h]		
	\centering
	\includegraphics[width=8cm]{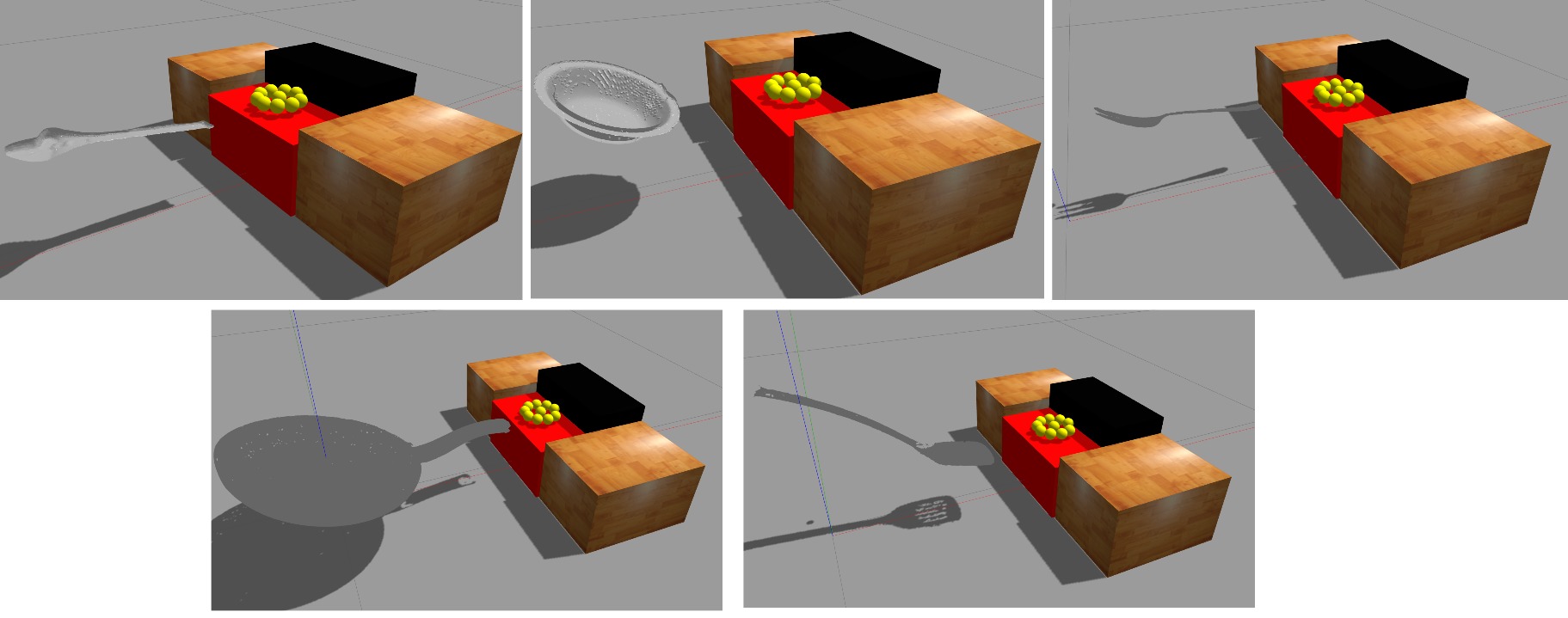}	
	\caption{Qualitative results of best projection for the lifitng pancake task. From right to left, top to bottom: adjustable spanner, child bowl, large fork, frying pan and a mesh spatula. Each subimage contains the best tool-pose found by the system for that tool.}
	\label{fig:lifting_projection_5_tools}	
\end{figure}


In Fig.~\ref{fig:lifting_projection_5_tools} we show projection of the lifting pancake task onto 5 different tools from ToolArtec: adjustable spanner, child bowl, large fork, frying pan and mesh spatula. Each subimage represents the best tool-pose found by the system for that tool for lifting pancake. We can see that the system finds creative uses by inverting the adjustable spanner and large fork in other to better lift. Even if lifting with the adjustable spanner does not lead to a good result, it is still the best tool-pose found by the system. This last statement is important because in our work we are interested in finding the best way of abstracting the input, but the input also restricts the best possible result since there might not be any abstraction that leads to a good result. Projection happens within an interaction between a concept and the raw input.

%

\subsubsection{Segmentation}
During segmentation \refabsecp{automatic_segmentation}, the running time for generating the $25$ segmentation options for a hammer mesh from ToolWeb with  $33182$ vertices and $63605$ faces is about $26$ seconds. We need to calculate the SDF values (see Sec.~\ref{sec:automatic_segmentation}) only once: this takes around $10$ seconds. Then it takes around $0.64$ to get the segments for each option.
Filtering the segmentation options to arrive at the best one takes the time required to fit a superquadric to each of the segments \refabsecp{exp_sq_fitting}.
\begin{figure}[h]		
	\centering
	\includegraphics[width=8cm]{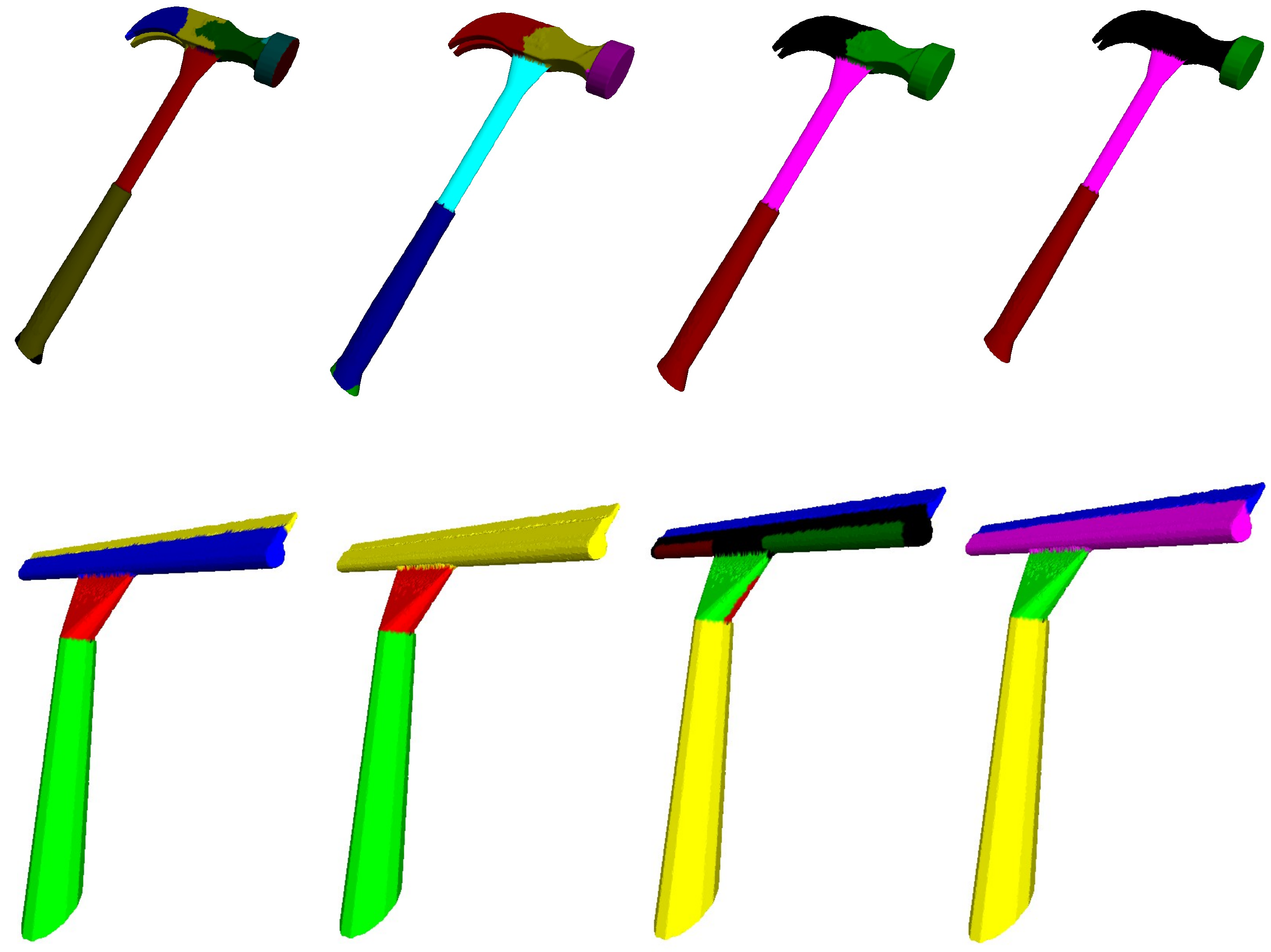}	
	\caption{Various segmentation options for hammer and squeegee from ToolWeb: different colours represent different segments. For the hammer, left to right, we have a fixed number of clusters $k=5$ and the smoothing parameter $\lambda$ varying from $\lambda=0.1$ to $\lambda=0.4$. For the squeegee, left to right, we have $(k=5$,$\lambda=0.4)$, $(k=5$,$\lambda=0.5)$, $(k=7$,$\lambda=0.2)$ and $(k=7$,$\lambda=0.3)$.}
	\label{fig:segm_options}	
\end{figure}

\section{Discussion and Future Work}
\label{sec:discussion}

In this section we first present the limitations we identified in our work, followed by suggestion on how to tackle them  \refabsecp{limitations}. Secondly, we propose future research directions on top of working on the limitations \refabsecp{future_work}. 

\subsection{Limitations}
\label{sec:limitations}

We identify four major shortcomings in our approach:\\
\begin{enumerate} 
\item \textit{Grasping}. Although we output a geometrical model as the grasping part, it does not have enough information for use in a real robot, since: (1) certain grasps, even after knowing where to grasp, still require planning to be able to bring the tool to the right position and orientation before the actual grasping; (2) the moment of inertia of an object can lead to an unstable grasp; and (3) the shape and scale of the grasping part might just be too different from the robotic hand to succeed.

\item \textit{Embodiment}. We do not evaluate on a real robot and our simulations are not embodied in any specific robot. This is the most important limitation since most works in this area are embodied in real robots. It could be that a given tool is not graspable by a specific robot hand or that, for instance, the tool is too heavy to be lifted.

\item \textit{Object Material}. We do not have representation of object material (this is clear in that our system misclassified all mugs as good for hammering because it does not represent the brittleness of porcelain). Also, we have no representation of elasticity or flexibility that can be useful to determine affordance for specific tasks. 

\item \textit{Tool articulation}. We do not have representation of tool articulation (moving parts in a tool, e.g., pliers), which can affect what happens when the tool is grasped one way or another or when it is used in a particular orientation. Additionally, no representation of more than two parts. Nonetheless, our tool representation could be extended to account for more parts, but integrating movable, linked or flexible parts would require more investigation in the required modifications. 
\end{enumerate}

Regarding the grasping limitation (1), we are still far from what would need to be done to integrate our framework into a real robot. However, even if incomplete we still believe our system could serve as a module inside other frameworks. Since certain grasps require planning, the system that incorporates our system will have to be able to plan given a certain grasp shape and scale: one system capable of this is the work by \cite{Tenorth2013KnowRob} and the integration with our system should be challenging, but not impossible. The amount of possible grasps that humans are capable of is immense, leading to a complex taxonomy of possible hand positions and finger pressure that is closely tied to the specific human hand anatomy \citep{Balasubramanian2014,Cutkosky1989}. Therefore, a complete grasping output would have to take into account the robotic hand, its grippers and torque output for the different parts. This could be achieved by training our system on a specific robotic and learning a function that predicts success given a grasping representation (with surface material, moment of inertia). This function could then act as a top-down pressure to guide the system towards the best grasping, balancing the choice between action and grasping part to achieve the best outcome. Finally, we could also incorporate a function that heavily penalises grasping parts that have its scale and shape parameters distant to ideal regions in the scale-shape space: these regions being the ones where the robot hand is most successful at grasping.

The embodiment shortcoming (2) can be tackled by looking for ways to integrate it into existing robot systems that allow for different modules, such as the work by \cite{Tenorth2013KnowRob} and investigsation into other possible collaboration with different robots in other research teams. The idea is to replicate all, or part of, our five tasks and see how well our system would cope with helping the robot assessing a new tool. However, it is important to note that we use a standard robotics simulator Gazebo that has integration with the robotic control software ROS, therefore it could be possible to integrate our system by tightly integrating the simulation with a specific robot.

As for the object material shortcoming (3), it would be interesting to gather a diverse dataset with tools with different materials and extend our system to be able to deal with them. A first crude representation of object material could be done by devising a numerical representation of brittleness, or flexibility and elasticity. However, the biggest challenge is: making a real robot perceive the materials from visual inspection or probing; and simulating those materials in order to learn a large dataset of how they affect a given task. One possible idea is that flexible parts could be seen as non-rigid shapes and a lot of work has been done in this area \citep{XieDeepShape2017,Pickup2015,Reuter2006}. The area has been increasing, evidenced by work such as \cite{Lian2015} that tries to benchmark all the different approaches. It might be possible to take advantage of some of those works and incorporate them into our work, but this would require further investigation.

Finally, concerning the tool articulation limitation (4), we can only deal with a one or two-part tool representaion. This is a complex problem and involves learning and dealing with different kinematic models and representations \citep{Khatib2014,Pillai2014,Katz2013}. One possible way of incorporating articulation in our work would require augmenting our tool representation to be a graph of parts and representing joints between those parts, considerly increasing the number of dimensions to represent a tool. We would also need to account for how to position the tool in the simulation considering that its parts may move very differently as we rotate and position the tool. Also, this is asusming a fixed number of parts, which can lead to very sparse (or redundant) vectors for tools composed of only one part (e.g. bowl). In our current representation the p-tool for a bowl would contain the same sub-vector for the grasping and the action part. This redudancy (or sparsity if we give all zeros to the rest of the p-tool) would increase as we extend our representation with more parts and with joints.

\subsection{Future Work}
\label{sec:future_work}

Additionally to working on solving the shortcomings presented in \refabsec{limitations}, we identify four future directions for our approach:

\begin{enumerate}[label=(\alph*)] 
\item \textit{Complex shapes}. It would be useful to extend our superquadric/superparaboloid representation to more generic surfaces although this should come at a price for fitting the surface to the data. This is an active area of research and it would require investigation into the balance and benefits to be found between a flexible representation and fitting optimisation.
\item \textit{Top-down influence}. Currently we plant seeds and fit options into sensor data to only then select the best option. We could call this a ``Darwinian'' approach that can be contrasted with a ``Lamarckian'' approach where the task function could influence the search for ways of abstracting sensor data. Instead of planting seeds we could think of starting at maxima values on the task function and try to fit those maxima onto the data and ``walk'' along points (i.e. p-tools) in the task function space and at the same time try to fit those points onto the data. This could be cast as an online optimisation problem that tries to maximise both the fitting score and the task function value.

\item \textit{Attention}. Our approach has no representation of attention, which is arguably an important part of human cognition \citep{Corbetta2002,PashlerMITAttention1998,DesimoneAttention1995} and A.I. systems, particularly for computer vision systems \citep{Sun2003}. Attention is valuable for explaining how primates can arrive at very accurate assessments with limited brain and time resources \citep{PashlerMITAttention1998}. Attention was also mentioned as crucial in research that is specifically about analogy-making in work by \cite[Sec. 8.2.]{Hofstadter1995a}. An attention mechanism would move our approach away from the random planting of seeds and possibly into a more efficient exploration of the search space. This relates closely to future direction (b) as attention is as a top-down process in itself. For instance, it has been shown that we have limited capacity for information and that we need selectivity to rule out unwanted information \cite{DesimoneAttention1995}. We could develop a mechanism to jump from one p-tool to another as the system is influenced by one or other aspect or assigning high probability for exploration of certain regions of the space (e.g. when searching for a tool to scoop grains we might have a strong belief that it needs to be a paraboloid, which is half of the space). Nonetheless, some room (even if small) for changing attention is essential to allow the agent to consider more strange possibilities. Attention is supposed to improve efficiency, but should not severely restrict the agent's ability to explore the space.

\item \textit{Principled projection}. It could be valuable to extend the work in this dissertation to a more principled approach. This could bring together future work directions (b) and (c) into a single framework. We want different features to exert different pressures for the final results: for one task, for instance, material can be very important (e.g. hammering a nail is not possible with brittle material); for another task, the most important aspect can be size of one or more dimensions (e.g. retrieving something in a gap); this is inspired by work from \cite{Hofstadter1995a} on analogy-making. The principled approach could take the form of a Bayesian model that assigns probabilities to exploration of regions of the p-tool space or, more complexly, probabilities to a pair of region and resources to be used for exploring that region. Also, inspired by \cite{LakeBBS2017}, it could also be a system that constructs programs that are recipes for assessing tools for a given task: the program could be a combination of primitive transformations to be done to a point cloud (e.g. segmenting, fitting, rotating, translating). Each task might result in more efficient programs for it: in order to hammer a nail, first check the tool's weight, if it is below a certain threshold do not even consider it; or including a guided segmentation to quickly find a narrow blade on the object to see if it is a good candidate for cutting lasagne.

\item \textit{Beyond tools}. It would be fruitful if we could extend our approach to go beyond tools and deal with manipulation tasks where the robot is interacting with objects without tools.  Tool-use in service robotics is the specific area in which we chose to show our projection mechanism, but we believe that its general principles could help in other manipulation tasks. 
\end{enumerate}

\section{Conclusion}


In this paper we have proposed an approach for affordance assessment of tools for everyday tasks in service robotics. We assess not only a point cloud's affordance score for a task, but also cues for manipulation: geometric models representing the grasping and end effector parts that enable positioning and orienting the tool for the task. Our approach introduces a tool model called p-tool that is able to represent a wide variety of sizes, shapes and grasp to end effector relationships. We are able to map from a point cloud to many p-tools; each p-tool represents an interpretation, or a usage of the point cloud. We use a synthetic dataset gathered from the web, and simulation, to learn task functions for different tasks mapping from a p-tool to an affordance score. At run-time, once the point cloud is mapped to a p-tool, we are able to assess its affordance score using the task function and also position and orient it for the task. 

The key point of our approach is that it is able to interpret a point cloud in many ways and is not limited by notions of ground truth category of objects. We wish to capture the idea that an object has many ways in which it can be interpreted as useful for a given goal. Bottom-up pressures from below and top-down pressures form above should be able to search through the possibilities and find a good trade-off in the interpretation being truthful to the data at the same time that is useful to a goal.


We evaluated on two point cloud datasets: the first scanned with Artec Eva 3D and the second with Kinect. Our results are compared with \cite{SchoelerTCDS2016}, the closest approach in the literature. We achieve better results while also outputting cues for manipulation and taking into account physical properties of the tool, such as mass and inertia. Furthermore, our system was much more resilient when tested on Kinect over Artec, actually improving with Kinect going from $40\%$ to $47\%$,   while the accuracy of \cite{SchoelerTCDS2016} went down from $46\%$ to $15\%$.

Finally, we highlight three shortcomings of our approach:
\begin{itemize}
\item It does not have prior knowledge over what is a good grasp, making its decision of where to grasp rely only on the task function
\item It is limited by the learned task function and it is not clear how that would generalise to a real robot in a real setting
\item  It was not tried on RGB-D data - in future work we intend on assess the system on RGB-D datasets
\end{itemize}

The work of this dissertation borrowed ideas from studies in metaphor and analogy-making in cognitive science, particularly the work by \cite{BipinBook1992}. While most works in this area apply to very high-level description models or toy-domains, we were able to successfully bring the projection idea into a real-world 3D vision domain for service robotics. Additionally, this dissertation demonstrated a pathway to semi-automatic learning from a small dataset and flexible assessment of tool affordances in everyday tool-use scenarios for service robots.

In \refabsec{system} we presented our fast sampling method that is useful for both selecting geometric fits as well as for rendering back a tool from its vector representation to a point cloud for simulation. This method is essential for being able to generate large datasets of tools for simulation from a very compact representation. In \refabsec{exp_ablation} this paper we presented an ablation study that showed how our system improves with the number of seeds and how a balanced weight between the fit score and the task function lead to overall better results: $91\%$ compared to $85\%$. This corroborates our idea that a concept should be both truthful to the data (fitting score) and also to the goal at hand  (task function value).

When bringing the cognitive science projection idea into a real-world domain we learned a few lessons. First, learning task function is really important since it allows us to select among alternatives in a more grounded way. In the work by \cite{BipinBook1992} there is no learning of a function capable of choosing what is the ``best'' metaphor among all the possible ones; also, in the seminal work on analogy by \cite{Hofstadter1995a}, the authors hand-code in an implicit way what is the stopping point for deciding on an analogical mapping (the final \textit{temperature} of the system). However, this temperature is not learned from data, but rather fine-tuned for the string-letter domain.  From this it seems that a lot of the focus is on endowing systems with flexibility, but with less focus on how to guide that flexibility in an explicit way to arrive at a best option. 

We are inspired by the human ability to change representation and believe that A.I. would benefit immensely by more work on this. Nonetheless, we also believe that working in real-world domains is essential to avoid the pitfalls of working on simple domains (specially the ones invented by the researchers themselves). Real-world domains are partially defined by having high-dimensional data, making the space of possible representations huge. Therefore, we chose to have a representation such that any datum (point cloud) could be fairly quickly abstracted into a lower-dimensional space (p-tool). And, importantly, that we could also go back from this representation to a lower-level datum. This is essential for approaches that want to be able to simulate, learn and imagine what would happen under different situations.

If we were to go to a different domain, such as language, we believe the same benefits should apply. Take a concept such as `meander': it could be used in its literal meaning of `a turn or winding of a stream' as in \textit{The meander eventually became isolated from the main stream} \footnote{https://www.merriam-webster.com/dictionary/meander}. However, we could also construct the sentence: \textit{The speaker was meandering during that talk}. As humans, when we listen to this for the first time we can understand what one meant by the metaphorical sentence; not only that, but we can also relate the meandering concept to actual moments that happen during the talk, for instance, with the speaker's personality or with the structure of the presentation. Both representations of our literal concept of `meander' and of our memory of the talk have to be changed in order for the mapping to be possible. Somehow we can take our memory of the talk and represent it in a space where the mapping from the literal meaning of `meander' can ``make sense''. Making sense is a very important part and it relates to having some evaluation function (e.g. the task function in our case) that can work in that space and define what are satisfactory mappings. Without such functions our brains would be lost in the huge space of possible representations and mappings. This points to the importance of having a `Lamarckian' approach where evaluation functions actually guide the representational switches. It remains a difficult area of research, though, specifying a way in which such system could be built, specially taking into account the speed at which humans are able to do such switches and evaluations.

In a broader perspective, we believe the epistemological stance of constructivism and research into changing representations have a lot to offer to A.I. That is, seeing a concept as not a one-pass abstraction and assessment from data, but as an interplay between the agent's goals and the low-level sensor data.

\begin{acks}
Thanks to the University of Aberdeen’s ABVenture Zone for equipment use.
We got a lot of very helpful advice and assistance from the following people (the first four did actual coding work on the project): Nikola Petkov, Krasimir Georgiev, Benjamin Nougier, Severin Fichtl, Subramanian Ramamoorthy, Michael Beetz, Andrei Haidu, John Alexander, Markus Schoeler, Nicolas Pugeault, David Cruickshank, Michael Chung and Naveed Khan
\end{acks}

\begin{funding}
Paulo Abelha is on a PhD studentship supported by the Brazilian agency CAPES through the program Science without Borders.
Frank Guerin received no specific grant from any funding agency in the public, commercial, or not-for-profit sectors. 
\end{funding}

\begin{dci}
The Authors declare that there is no conflict of interest
\end{dci}

\bibliographystyle{SageH}
\bibliography{library2}

\begin{thebibliography}{57}
\providecommand{\natexlab}[1]{#1}
\providecommand{\url}{}
\providecommand{\urlprefix}{URL }
\expandafter\ifx\csname urlstyle\endcsname\relax
  \providecommand{\doi}[1]{DOI:\discretionary{}{}{}#1}\else
  \providecommand{\doi}{DOI:\discretionary{}{}{}\begingroup
  \urlstyle{rm}\Url}\fi

\bibitem[{Abelha(2017)}]{AbelhaSQ2017}
Abelha P (2017) {Close-to-uniform Sampling of Superquadric Point Clouds with
  Normals (under review)}.

\bibitem[{Abelha and Guerin(2017)}]{AbelhaIROS2017}
Abelha P and Guerin F (2017) {Learning How a Tool Affords by Simulating 3D
  Models from the Web}.
\newblock In: \emph{2017 IEEE/RSJ International Conference on Intelligent
  Robots and Systems (IROS)}. Vancouver, p. (to be published).

\bibitem[{Abelha et~al.(2016)Abelha, Guerin and Schoeler}]{AbelhaICRA2016}
Abelha P, Guerin F and Schoeler M (2016) {A model-based approach to finding
  substitute tools in 3D vision data}.
\newblock In: \emph{2016 IEEE International Conference on Robotics and
  Automation (ICRA)}. IEEE.
\newblock ISBN 978-1-4673-8026-3, pp. 2471--2478.

\bibitem[{Agostini et~al.(2015)Agostini, {Javad Aein}, Szedmak, Aksoy, Piater
  and Worgotter}]{AgostiniIROS2015}
Agostini A, {Javad Aein} M, Szedmak S, Aksoy EE, Piater J and Worgotter F
  (2015) {Using structural bootstrapping for object substitution in robotic
  executions of human-like manipulation tasks}.
\newblock In: \emph{2015 IEEE/RSJ International Conference on Intelligent
  Robots and Systems (IROS)}, volume 2015-Decem. IEEE.
\newblock ISBN 978-1-4799-9994-1, pp. 6479--6486.

\bibitem[{Balasubramanian and Santos(2014)}]{Balasubramanian2014}
Balasubramanian R and Santos VJ (2014) {The human hand as an inspiration for
  robot hand development}.
\newblock \emph{Springer Tracts in Advanced Robotics} 95: 219--246.

\bibitem[{Barr(1981)}]{Barr1981}
Barr AH (1981) {Superquadrics and Angle- Preserving Transformations}.
\newblock \emph{Computer Graphics and Applications, IEEE} 1(1): 11 -- 23.

\bibitem[{Bengio et~al.(2013)Bengio, Courville and
  Vincent}]{BengioRepLearn2015}
Bengio Y, Courville A and Vincent P (2013) {Representation Learning: A Review
  and New Perspectives}.
\newblock \emph{IEEE Transactions on Pattern Analysis and Machine Intelligence}
  35(8): 1798--1828.

\bibitem[{Biegelbauer et~al.(2008)Biegelbauer, Vincze and
  Wohlkinger}]{Biegelbauer2008}
Biegelbauer G, Vincze M and Wohlkinger W (2008) {Model-based 3D object
  detection}.
\newblock \emph{Machine Vision and Applications} 21(4): 497--516.

\bibitem[{Bu et~al.(2014)Bu, Liu, Han, Wu and Ji}]{Bu2014}
Bu S, Liu Z, Han J, Wu J and Ji R (2014) {Learning High-Level Feature by Deep
  Belief Networks for 3-D Model Retrieval and Recognition}.
\newblock \emph{IEEE Transactions on Multimedia} 16(8): 2154--2167.

\bibitem[{Chu et~al.(2016)Chu, Fitzgerald and Thomaz}]{Chu2016}
Chu V, Fitzgerald T and Thomaz AL (2016) {Learning Object Affordances by
  Leveraging the Combination of Human-Guidance and Self-Exploration}.
\newblock In: \emph{The Eleventh ACM/IEEE International Conference on Human
  Robot Interaction}. New Zeland: IEEE Press.
\newblock ISBN 9781467383707, pp. 221--228.

\bibitem[{Corbetta and Shulman(2002)}]{Corbetta2002}
Corbetta M and Shulman GL (2002) {Control of Goal-Directed and Stimulus-Driven
  Attention in the Brain}.
\newblock \emph{Nature Reviews Neuroscience} 3(3): 215--229.

\bibitem[{Cutkosky(1989)}]{Cutkosky1989}
Cutkosky MR (1989) {On Grasp Choice, Grasp Models, and the Design of Hands for
  Manufacturing Tasks}.
\newblock \emph{IEEE Transactions on Robotics and Automation} 5(3): 269--279.

\bibitem[{Davis and Marcus(2015)}]{DavisMarcus2015}
Davis E and Marcus G (2015) {Commonsense reasoning and commonsense knowledge in
  artificial intelligence}.
\newblock \emph{Communications of the ACM} 58(9): 92--103.

\bibitem[{Dehban et~al.(2016)Dehban, Jamone, Kampff and
  Santos-Victor}]{DehbanICRA2016}
Dehban A, Jamone L, Kampff AR and Santos-Victor J (2016) {Denoising
  auto-encoders for learning of objects and tools affordances in continuous
  space}.
\newblock In: \emph{2016 IEEE International Conference on Robotics and
  Automation (ICRA)}, volume 2016-June. IEEE.
\newblock ISBN 978-1-4673-8026-3, pp. 4866--4871.

\bibitem[{Desimone and Duncan(1995)}]{DesimoneAttention1995}
Desimone R and Duncan J (1995) {Neural Mechanisms of Selective Visual}.
\newblock \emph{Annual Review of Neuroscience} 18(1): 193--222.

\bibitem[{Ersen et~al.(2017)Ersen, Oztop and Sariel}]{Ersen2017}
Ersen M, Oztop E and Sariel S (2017) {Cognition-Enabled Robot Manipulation in
  Human Environments: Requirements, Recent Work, and Open Problems}.
\newblock \emph{IEEE Robotics {\&} Automation Magazine} 24(3): 108--122.

\bibitem[{{Feng Han} and {Song-Chun Zhu}(2009)}]{HanTPAMI2009}
{Feng Han} and {Song-Chun Zhu} (2009) {Bottom-Up/Top-Down Image Parsing with
  Attribute Grammar}.
\newblock \emph{IEEE Transactions on Pattern Analysis and Machine Intelligence}
  31(1): 59--73.

\bibitem[{Harel et~al.(2014)Harel, Kravitz and Baker}]{Harel2014a}
Harel A, Kravitz DJ and Baker CI (2014) {Task context impacts visual object
  processing differentially across the cortex}.
\newblock \emph{Proceedings of the National Academy of Sciences} 111(10):
  E962--E971.

\bibitem[{Hofstadter and Mitchell(1995)}]{Hofstadter1995a}
Hofstadter D and Mitchell M (1995) {The Copycat Project : A Model of Mental
  Fluidity and}.
\newblock In: \emph{Fluid concepts and creative analogies: computer models of
  the fundamental mechanisms of thought.}, 1995 edition, chapter~5. London:
  Penguim Books.
\newblock ISBN 978-0140258356, pp. 205--267.

\bibitem[{Indurkhya(1992)}]{BipinBook1992}
Indurkhya B (1992) \emph{{Metaphor and Cognition}}.
\newblock Dordrecht, The Netherlands: Kluwer Academic Publishers.

\bibitem[{Jakli{\v{c}} et~al.(2000)Jakli{\v{c}}, Leonardis and
  Solina}]{JacklicSQBook}
Jakli{\v{c}} A, Leonardis A and Solina F (2000) \emph{{Segmentation and
  Recovery of Superquadrics}}, \emph{Computational Imaging and Vision},
  volume~20.
\newblock Dordrecht: Springer Netherlands.

\bibitem[{Jamone et~al.(2016)Jamone, Ugur, Cangelosi, Fadiga, Bernardino,
  Piater and Santos-Victor}]{JamoneTCDS2016}
Jamone L, Ugur E, Cangelosi A, Fadiga L, Bernardino A, Piater J and
  Santos-Victor J (2016) {Affordances in psychology, neuroscience and robotics:
  a survey}.
\newblock \emph{IEEE Transactions on Cognitive and Developmental Systems}
  PP(99): 1.

\bibitem[{Katz et~al.(2013)Katz, Kazemi, {Andrew Bagnell} and
  Stentz}]{Katz2013}
Katz D, Kazemi M, {Andrew Bagnell} J and Stentz A (2013) {Interactive
  segmentation, tracking, and kinematic modeling of unknown 3D articulated
  objects}.
\newblock \emph{Proceedings - IEEE International Conference on Robotics and
  Automation} : 5003--5010.

\bibitem[{Khatib et~al.(2014)Khatib, Kumar and Sukhatme}]{Khatib2014}
Khatib O, Kumar V and Sukhatme G (2014) {Experimental Robotics: The 12th
  International Symposium on Experimental Robotics ABC}.
\newblock \emph{Springer Tracts in Advanced Robotics} 79: 301--315.

\bibitem[{Koppula et~al.(2012)Koppula, Gupta and Saxena}]{Koppula2012}
Koppula HS, Gupta R and Saxena A (2012) {Learning Human Activities and Object
  Affordances from RGB-D Videos}.
\newblock \emph{The International Journal of Robotics Research} 32(8):
  951--970.

\bibitem[{Kraft et~al.(2010)Kraft, Detry, Pugeault, Baseski, Guerin, Piater and
  Kruger}]{KraftTCDS2010}
Kraft D, Detry R, Pugeault N, Baseski E, Guerin F, Piater JH and Kruger N
  (2010) {Development of Object and Grasping Knowledge by Robot Exploration}.
\newblock \emph{IEEE Transactions on Autonomous Mental Development} 2(4):
  368--383.

\bibitem[{Krainin et~al.(2011)Krainin, Henry, Ren and Fox}]{KraininIJRR2011}
Krainin M, Henry P, Ren X and Fox D (2011) {Manipulator and object tracking for
  in-hand 3D object modeling}.
\newblock \emph{The International Journal of Robotics Research} 30(11):
  1311--1327.

\bibitem[{Kroemer et~al.(2012)Kroemer, Ugur, Oztop and
  Peters}]{KroemerICRA2012}
Kroemer O, Ugur E, Oztop E and Peters J (2012) {A kernel-based approach to
  direct action perception}.
\newblock In: \emph{2012 IEEE International Conference on Robotics and
  Automation}. IEEE.
\newblock ISBN 978-1-4673-1405-3, pp. 2605--2610.

\bibitem[{Lake et~al.(2016)Lake, Ullman, Tenenbaum and Gershman}]{LakeBBS2017}
Lake BM, Ullman TD, Tenenbaum JB and Gershman SJ (2016) {Building Machines That
  Learn and Think Like People}.
\newblock \emph{Behavioral and Brain Sciences} (May 2017): 1--101.

\bibitem[{LeCun et~al.(2015)LeCun, Bengio, Hinton, Y., Y. and G.}]{LeCun2015}
LeCun Y, Bengio Y, Hinton G, Y L, Y B and G H (2015) {Deep learning}.
\newblock \emph{Nature} 521(7553): 436--444.

\bibitem[{Lenz et~al.(2015)Lenz, Lee and Saxena}]{LenzIJRR2015}
Lenz I, Lee H and Saxena A (2015) {Deep learning for detecting robotic grasps}.
\newblock \emph{The International Journal of Robotics Research} 34(4-5):
  705--724.

\bibitem[{Levine et~al.(2015)Levine, Finn, Darrell and Abbeel}]{Levine2015}
Levine S, Finn C, Darrell T and Abbeel P (2015) {End-to-End Training of Deep
  Visuomotor Policies}.
\newblock \emph{Journal of Machine Learning Research} 17: 1--40.

\bibitem[{Lian et~al.(2015)Lian, Zhang, Choi, Elnaghy, Furuya, Giachetti and
  Guler}]{Lian2015}
Lian Z, Zhang J, Choi S, Elnaghy H, Furuya T, Giachetti A and Guler RA (2015)
  {SHREC'15 Track: Non-rigid 3D Shape Retrieval}.
\newblock \emph{EurographicsWorkshop on 3D Object Retrieval} .

\bibitem[{Mahler et~al.(2017)Mahler, Liang, Niyaz, Laskey, Doan, Liu, Ojea and
  Goldberg}]{MahlerDexNet2017}
Mahler J, Liang J, Niyaz S, Laskey M, Doan R, Liu X, Ojea JA and Goldberg K
  (2017) {Dex-Net 2.0: Deep Learning to Plan Robust Grasps with Synthetic Point
  Clouds and Analytic Grasp Metrics} .

\bibitem[{Mar et~al.(2015)Mar, Tikhanoff, Metta and Natale}]{MarICHR2015}
Mar T, Tikhanoff V, Metta G and Natale L (2015) {Multi-model approach based on
  3D functional features for tool affordance learning in robotics}.
\newblock In: \emph{2015 IEEE-RAS 15th International Conference on Humanoid
  Robots (Humanoids)}, volume 270273. IEEE.
\newblock ISBN 978-1-4799-6885-5, pp. 482--489.

\bibitem[{Myers et~al.(2015)Myers, Teo, Fermuller and
  Aloimonos}]{MyersICRA2015}
Myers A, Teo CL, Fermuller C and Aloimonos Y (2015) {Affordance detection of
  tool parts from geometric features}.
\newblock In: \emph{2015 IEEE International Conference on Robotics and
  Automation (ICRA)}. IEEE.
\newblock ISBN 978-1-4799-6923-4, pp. 1374--1381.

\bibitem[{Pashler and Sutherland(1998)}]{PashlerMITAttention1998}
Pashler HE and Sutherland S (1998) \emph{{The psychology of attention}},
  volume~15.
\newblock MIT press Cambridge, MA.

\bibitem[{Pickup et~al.(2015)Pickup, Sun, Rosin and Martin}]{Pickup2015}
Pickup D, Sun X, Rosin PL and Martin RR (2015) {Euclidean-distance-based
  canonical forms for non-rigid 3D shape retrieval}.
\newblock \emph{Pattern Recognition} 48(8): 2500--2512.

\bibitem[{Pillai et~al.(2014)Pillai, Walter and Teller}]{Pillai2014}
Pillai S, Walter MR and Teller S (2014) {Learning Articulated Motions From
  Visual Demonstration}.
\newblock \emph{Robotics: Science and Systems} .

\bibitem[{Rasmussen and Williams(2006)}]{RasmussenGPBook}
Rasmussen CE and Williams CK (2006) \emph{{Gaussian Processes for Machine
  Learning}}, volume~1.
\newblock Cambridge: MIT Press.

\bibitem[{Reuter et~al.(2006)Reuter, Wolter and Peinecke}]{Reuter2006}
Reuter M, Wolter FE and Peinecke N (2006) {Laplace-Beltrami spectra as
  'Shape-DNA' of surfaces and solids}.
\newblock \emph{CAD Computer Aided Design} 38(4): 342--366.

\bibitem[{{\v{S}}ahin et~al.(2007){\v{S}}ahin, {\c{C}}akmak, Do{\"{u}}gar,
  {\"{U}}gur and {\"{U}}{\c{c}}oluk}]{Sahin2007}
{\v{S}}ahin E, {\c{C}}akmak M, Do{\"{u}}gar MR, {\"{U}}gur E and
  {\"{U}}{\c{c}}oluk G (2007) {To afford or not to afford: A new formalization
  of affordances toward affordance-based robot control}.
\newblock \emph{Adaptive Behavior} 15(4): 447--472.

\bibitem[{Savva et~al.(2015)Savva, Chang and Hanrahan}]{Savva2015}
Savva M, Chang AX and Hanrahan P (2015) {Semantically-enriched 3D models for
  common-sense knowledge}.
\newblock In: \emph{2015 IEEE Conference on Computer Vision and Pattern
  Recognition Workshops (CVPRW)}, volume 2015-Octob. IEEE.
\newblock ISBN 978-1-4673-6759-2, pp. 24--31.

\bibitem[{Schmidhuber(2015)}]{Schmidhuber2015}
Schmidhuber J (2015) {Deep Learning in neural networks: An overview}.
\newblock \emph{Neural Networks} 61(7553): 85--117.

\bibitem[{Schoeler and W{\"{o}}rg{\"{o}}tter(2016)}]{SchoelerTCDS2016}
Schoeler M and W{\"{o}}rg{\"{o}}tter F (2016) {Bootstrapping the Semantics of
  Tools: Affordance analysis of real world objects on a per-part basis}.
\newblock \emph{IEEE Trans. on Autonomous Mental Development} 8(2): 84--98.

\bibitem[{Shapira et~al.(2008)Shapira, Shamir and Cohen-Or}]{ShapiraSDF2008}
Shapira L, Shamir A and Cohen-Or D (2008) {Consistent mesh partitioning and
  skeletonisation using the shape diameter function}.
\newblock \emph{Visual Computer} 24(4): 249--259.

\bibitem[{Sinapov and Stoytchev(2008)}]{SinapovICDL2008}
Sinapov J and Stoytchev A (2008) {Detecting the functional similarities between
  tools using a hierarchical representation of outcomes}.
\newblock In: \emph{2008 7th IEEE International Conference on Development and
  Learning}. IEEE.
\newblock ISBN 978-1-4244-2661-4, pp. 91--96.

\bibitem[{Sun and Fisher(2003)}]{Sun2003}
Sun Y and Fisher R (2003) {Object-based visual attention for computer vision}.
\newblock \emph{Artificial Intelligence} 146(1): 77--123.

\bibitem[{Tenorth and Beetz(2013)}]{Tenorth2013KnowRob}
Tenorth M and Beetz M (2013) {KnowRob: A knowledge processing infrastructure
  for cognition-enabled robots}.
\newblock \emph{The International Journal of Robotics Research} 32(5):
  566--590.

\bibitem[{Tenorth et~al.(2013)Tenorth, Profanter, Balint-Benczedi and
  Beetz}]{TenorthIROS2013}
Tenorth M, Profanter S, Balint-Benczedi F and Beetz M (2013) {Decomposing CAD
  models of objects of daily use and reasoning about their functional parts}.
\newblock In: \emph{2013 IEEE/RSJ International Conference on Intelligent
  Robots and Systems}. IEEE.
\newblock ISBN 978-1-4673-6358-7, pp. 5943--5949.

\bibitem[{Ugur et~al.(2011)Ugur, Celikkanat, Sahin, Nagai and
  Oztop}]{UgurCSNO2011}
Ugur E, Celikkanat H, Sahin E, Nagai Y and Oztop E (2011) {Learning to grasp
  with parental scaffolding}.
\newblock In: \emph{2011 11th IEEE-RAS International Conference on Humanoid
  Robots}. IEEE.
\newblock ISBN 978-1-61284-868-6, pp. 480--486.

\bibitem[{Varadarajan and Vincze(2014)}]{KarthikICPR2014}
Varadarajan KM and Vincze M (2014) {4-D Space-Time Mereotopogeometry-Part
  Connectivity Calculus for Visual Object Representation}.
\newblock In: \emph{Pattern Recognition (ICPR), 2014 22nd International
  Conference on}. pp. 4316--4321.

\bibitem[{Wang et~al.(2014)Wang, Hindriks and Babuska}]{Wang2014a}
Wang C, Hindriks KV and Babuska R (2014) {Effective transfer learning of
  affordances for household robots}.
\newblock In: \emph{4th International Conference on Development and Learning
  and on Epigenetic Robotics}, 1. Genoa: IEEE.
\newblock ISBN 978-1-4799-7540-2, pp. 469--475.

\bibitem[{Welke et~al.(2010)Welke, Issac, Schiebener, Asfour and
  Dillmann}]{WelkeICRA2010}
Welke K, Issac J, Schiebener D, Asfour T and Dillmann R (2010) {Autonomous
  acquisition of visual multi-view object representations for object
  recognition on a humanoid robot}.
\newblock In: \emph{Robotics and Automation (ICRA), 2010 IEEE International
  Conference on}. IEEE.
\newblock ISBN 978-1-4244-5038-1, pp. 2012--2019.

\bibitem[{Wu et~al.(2014)Wu, Sun, Long, Huang, Cohen-Or, Gong, Deussen and
  Chen}]{WuACMTG2014}
Wu S, Sun W, Long P, Huang H, Cohen-Or D, Gong M, Deussen O and Chen B (2014)
  {Quality-driven poisson-guided autoscanning}.
\newblock \emph{ACM Transactions on Graphics} 33(6): 1--12.

\bibitem[{Xie et~al.(2017)Xie, Dai, Zhu, Wong and Fang}]{XieDeepShape2017}
Xie J, Dai G, Zhu F, Wong EK and Fang Y (2017) {DeepShape: Deep-Learned Shape
  Descriptor for 3D Shape Retrieval}.
\newblock \emph{IEEE Transactions on Pattern Analysis and Machine Intelligence}
  39(7): 1335--1345.

\bibitem[{Yaz and Lorior(2012)}]{YazTSMSegmentation2012}
Yaz I and Lorior S (2012) {CGAL 4.10 - Triangulated Surface Mesh Segmentation.}

\end{thebibliography}

\end{document}